\documentclass{article}

     \PassOptionsToPackage{numbers, compress}{natbib}

\usepackage[preprint]{neurips_2025}

\usepackage[utf8]{inputenc} 
\usepackage[T1]{fontenc}    
\usepackage{hyperref}       
\usepackage{url}            
\usepackage{booktabs}       
\usepackage{amsfonts}       
\usepackage{nicefrac}       
\usepackage{microtype}      
\usepackage{xcolor}         

\usepackage{graphicx}       
\usepackage{wrapfig}        

\usepackage{multirow} 
\usepackage{colortbl}  
\usepackage{amsmath} 

\newcommand{\ourdataset}[1]{\textsc{GuessBench}}

\title{\ourdataset{}: Sensemaking Multimodal \\ Creativity in the Wild}

\author{Zifeng Zhu\thanks{equal contribution} \ \textsuperscript{1} \ Shangbin Feng\footnotemark[1] \ \textsuperscript{2} \ Herun Wan\textsuperscript{1} \ Ningnan Wang\textsuperscript{1} \ Minnan Luo\textsuperscript{1} \ Yulia Tsvetkov\textsuperscript{2} \\
\textsuperscript{1}Xi'an Jiaotong University \ \ \ \ \ \textsuperscript{2}University of Washington \\
\texttt{zivenzhu@stu.xjtu.edu.cn} \ \ \ \texttt{shangbin@cs.washington.edu}
}

\begin{document}

\definecolor{Gray}{gray}{0.93} 
\newcolumntype{a}{>{\columncolor{Gray}}r} 

\maketitle

\begin{abstract}

We propose \ourdataset{}, a novel benchmark that evaluates Vision Language Models (VLMs) on modeling the pervasive, noisy, and pluralistic human creativity. \ourdataset{} sources data from ``Guess the Build'', an online multiplayer Minecraft minigame where one player constructs a Minecraft build given a concept (e.g., caterpillar) and others try to guess it with natural language hints, presenting a pristine testbed for sensemaking creativity in the wild with VLMs acting as guessers. We curate 1500 images from the actual gameplay and design 2000 problems spanning static and dynamic image settings, natural language hints of varying completeness, and more. Extensive experiments with six open/API VLMs and five reasoning enhancement approaches demonstrate that \ourdataset{} presents a uniquely challenging task in creativity modeling: even the start-of-the-art GPT-4o is incorrect on 34\% of instances, while we observe a huge performance gap (13.87\% vs. 53.93\% on average) between open and API models. When used as a resource to improve VLMs, fine-tuning on the reasoning traces for \ourdataset{} problems improves visual perception tasks by 15.36\% on average. Further analysis reveals that VLM performance in creativity sensemaking correlates with the frequency of the concept in training data, while the accuracy drops sharply for concepts in underrepresented cultural contexts and low-resource languages.
\end{abstract}

\section{Introduction}
\label{Introduction}

Vision Language Models (VLMs) have demonstrated remarkable capabilities across perception, knowledge, and reasoning problems~\cite{gpt4o, bai2025qwen2, wang2024enhancing, yao2024minicpm}. From objective tasks to subjective contexts, recent works have explored creative tasks for generative models, evaluating their skills in creative generation and problem-solving across language~\cite{shen2025societal, minhturning, wang2024create, lv-etal-2024-subjective}, image~\cite{fang2025creation, white2025collaborating, lifshitz2023steve, han2025enhancing, ng2024partcraft}, audio~\cite{cherep2024creative}, and video modalities~\cite{miller2024enhanced, han2024using, feng2024ccedit, jiang2024reinforcement, wang2024zola}.

We identify two key gaps in the research of creative generative modeling. 1) \emph{Creativity in the wild}: while existing research holds VLMs to high artistic standards~\cite{chakrabarty2024art, tang-etal-2024-creative, wang2023creative, zhang2024hidiffusion}, it struggles to incorporate the imperfect creativity of diverse VLM users. Their creativity is \emph{pervasive}, as many of their VLM requests (e.g., polishing a photo and generating a flowchart) require creative decision making \citep{kim2025bridging}; their creativity is \emph{noisy}, as the average VLM user is not artistically trained and did not memorize the elite art in VLM training data \citep{barton2013arts}; their creativity is \emph{pluralistic}, as different individuals could have varying interpretations for the same entity and concept \citep{sorensen2024position, feng2024modular}. As such, reflecting and modeling creativity \emph{in the wild} is a crucial step in aligning the creative capabilities of VLMs with diverse VLM users. 2) \emph{Sensemaking creativity}: while most research focuses on \emph{generative} creativity where models are tasked with generating engaging text or pretty images~\cite{lu2024ai, mc-bench, peng2025probing, gokaslan2024commoncanvas}, there is limited exploration on model capabilities in analyzing and decoding creative constructs. Quantifying and augmenting VLMs' skills in understanding and \emph{sensemaking} creativity could assist art teaching, build a reward model for creativity, and more.

\begin{figure}[t]
    \centering
    \includegraphics[width=\linewidth]{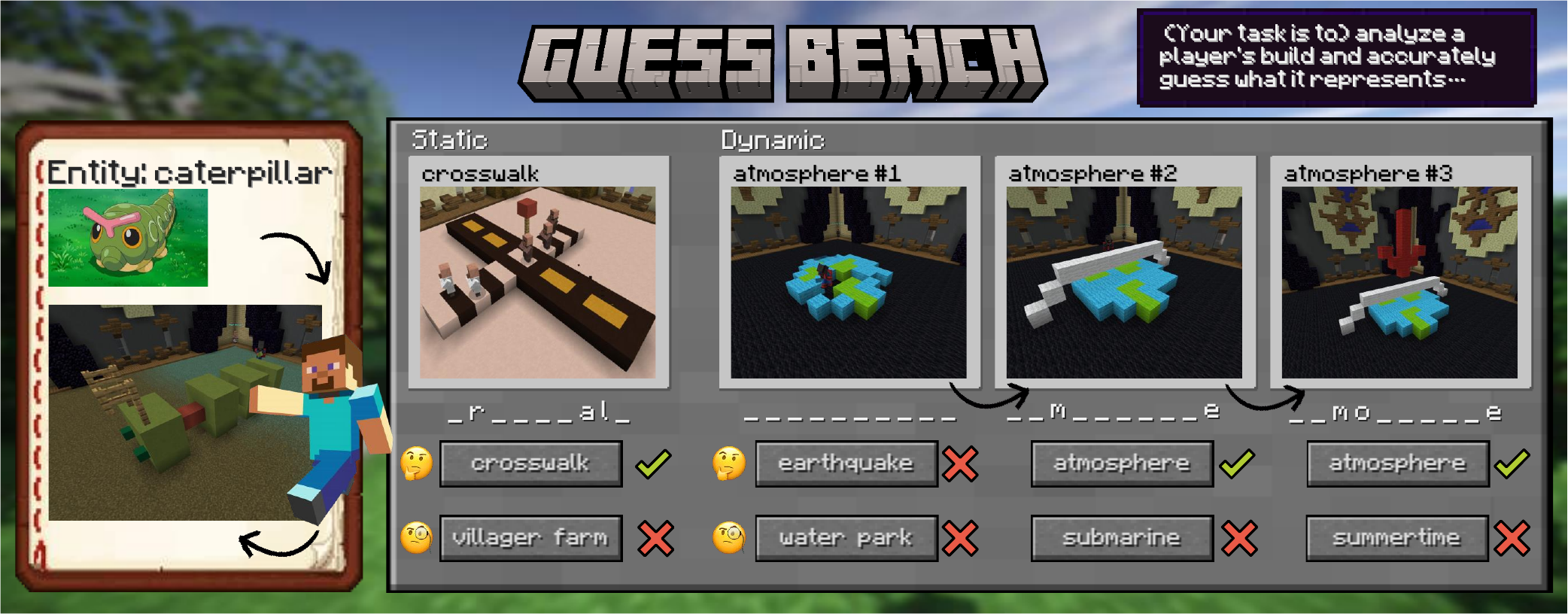}
    \caption{\ourdataset{} consists of player-constructed Minecraft builds representing real-world entities and concepts. In each problem, VLMs are required to infer what the build represents based on the provided image(s) and hint(s). We curate two settings: the static setting where only one image and hint are provided, and the dynamic setting where two sequences of progressively refined builds and corresponding hints are presented. In the dynamic setting, if the model answers correctly at any attempt, all subsequent attempts are considered correct.}
    \vspace{-10pt}
    \label{fig: overview}
\end{figure}

To this end, we propose \ourdataset{}, a creativity understanding dataset where VLMs guess and work out the underlying theme of a Minecraft image (Figure~\ref{fig: overview}). \ourdataset{} is sourced from ``Guess the Build'', a minigame on the Hypixel Minecraft multiplayer network\footnote{\url{https://hypixel.net}}: Given a concept (e.g., caterpillar or oasis), one player creates a Minecraft build conveying the concept, while other players try to guess the concept with optional natural language hints. We curate 1500 images from the actual gameplay and design 2000 problems under two evaluation settings: \emph{static}, where the VLM is only presented with the completed Minecraft build for one attempt, and \emph{dynamic}, where the VLM is presented with a sequence of images where the build is increasingly complete for multiple attempts. We posit that \ourdataset{} presents a pristine \emph{creativity in-the-wild} setting where data is sourced from a diverse and global player community with a spectrum of language and cultural backgrounds.

We evaluate a wide range of state-of-the-art VLMs and enhancement approaches (e.g., Self-Refine~\cite{madaan2023self} and Image Retrieval~\citep{zhu2024enhancing}) on \ourdataset{}. We find that \emph{creativity sensemaking} is challenging, even the state-of-the-art GPT-4o only achieves 57.8\% and 66.0\% in static and dynamic settings. \ourdataset{} also introduces a quantitatively novel~\cite{liautobencher} dataset: compared to various VQA datasets~\cite{lu2023mathvista, guan2024hallusionbench}, \ourdataset{} is more challenging, better separates model performance, and reveals novel insights about model strengths and weaknesses (Figure~\ref{fig: autobench_metrics}). \ourdataset{} could also be a useful training resource: by fine-tuning VLMs on the reasoning traces in \ourdataset{}, their performance on visual datasets improves by 15.36\% on average. Further analysis reveals that VLM creativity sensemaking degrades substantially for low-resource languages, correlates with the frequency of the concept in training data, and suffers from sycophancy when facing contradictory user requests.





\section{\ourdataset{}}

We propose \ourdataset{}, a novel and challenging benchmark designed to evaluate the creative understanding capabilities of VLMs. \ourdataset{} is based on the ``Guess the Build'' Minecraft minigame, where players construct a Minecraft build given a concept (e.g. caterpillar) and others try to guess it with natural language hints. We first introduce two problem settings for VLMs acting as a guesser, \emph{static} and \emph{dynamic}, where the Minecraft build image either stays fixed or gradually becomes more complete as the player develops their build (\S\ref{2.1: Task Setting}). To ensure the reasonableness, diversity, and difficulty of the problems, we collect images from real Minecraft gameplay scenarios and design well-crafted natural language hints with varying difficulty levels (\S\ref{2.2: Data Curation}). We then describe the evaluation metrics in \S\ref{2.3: Evaluation metric} and present dataset statistics in Table~\ref{tab:overall_statistics}.

\subsection{Task Setting}
\label{2.1: Task Setting}
\paragraph{Static Task Setting.} Given an image of a Minecraft build $b$ and a corresponding natural language hint $h$, the VLM's goal is to decipher player creativity and identify what concept does the build represent. This task can be formally defined as: $\mathrm{VLM}(b, h) = c$, where \(\mathrm{VLM}(\cdot)\) denotes the VLM, and $c$ is the textual output representing the VLM's guessed concept (e.g. oasis or caterpillar).

\paragraph{Dynamic Task Setting.} Consider a sequence of Minecraft build images denoted by $\mathbf{b}_T = (b_1, b_2, \ldots, b_T)$, where $b_t$ represents the build image at time step $t$. Correspondingly, we define a sequence of hints as $\mathbf{h}_T = (h_1, h_2, \ldots, h_T)$, and a sequence of the VLM’s previous guesses as $\mathbf{c}_{T-1} = (c_1, c_2, \ldots, c_{T-1})$, where $\mathbf{c_0}$ is set as an empty sequence to indicate no previous guesses at the start. In our proposed \ourdataset{}, both the Minecraft build images and natural language hints become progressively more complete through time, presenting a temporal and dynamic setting where VLMs need to work with incomplete information. Based on this setup, the task can be formally defined as: $\mathrm{VLM}(\mathbf{b}_T, \mathbf{h}_T, \mathbf{c}_{T-1}) = \mathbf{c}_T, T \geq 1$.

\begin{wraptable}[24]{r}{5.8cm}
\vspace{-3.5ex}
\scalebox{0.85}{
\begin{tabular}{lr}
\hline
\toprule
\textbf{Statistics}            & \textbf{Value}   \\
\midrule

Minecraft build sets             & $500$  
\\
\textit{Answer}   &                  \\
- Unique answers/tokens            & $424/678$     \\
- Maximum/Average length           & $4/2.1$       \\
\midrule

\textbf{Static Task}            &                  \\
\textit{Minecraft Build Images}            &                  \\
- Total Images                   & $500$
\\
- Images per set                 & $1$
\\
- Average size (px)              & $1188\times753$   \\
- Maximum size (px)              & $1920\times1080$ \\

\textit{Hints} &                  \\
- Unique hints/tokens            & $481/120$
\\
- Maximum/Average length         & $127/78.3$ 
\\

\midrule

\textbf{Dynamic Task}            &                  \\
\textit{Minecraft Build Images}            &                  \\
- Total images                   & $1500$  
\\
- Images per set                 & $3$
\\
- Average size (px)              & $1150\times721$   \\
- Maximum size (px)              & $1920\times1080$ \\

\textit{Hints} &                  \\
- Unique hints/tokens            & $933/125$
\\
- Maximum/Average length         & $127/61.2$ 
\\

\bottomrule
\hline
\end{tabular}}
\caption{Statistics of \ourdataset{}. The unique tokens and lengths of hints and answers are measured using the GPT-4o tokenizer.}
\label{tab:overall_statistics}
\end{wraptable}

\subsection{Data Curation} 
\label{2.2: Data Curation}

\paragraph{Build Collection.} To collect representative visual data, we manually participate in the ``Guess the Build'' online multiplayer game on the Hypixel server and capture screenshots of the constructed builds. During the image selection process, we also remove low-quality builds such as those where players spell out the answer using blocks instead of building its visual representation, or where no meaningful construction is present (e.g., merely holding an apple when the target word is ``apple''). In total, \ourdataset{} comprises of 500 carefully curated build sets. For the \emph{dynamic} task, we set the number of attempts \( T = 3 \), collecting three successive Minecraft build images for each set. For the \emph{static} task, we use the third-attempt image from the dynamic task (i.e., \( b_3 \)) as the representative build for each set. This results in a total of 1500 images featuring diverse concepts and varying levels of completeness.

\paragraph{Hint Generation.} To more faithfully simulate the complete gameplay of ``Guess the Build'', we draw inspiration from its progressive hint-revealing mechanism and design corresponding hints for each guessing attempt.

Specifically, suppose the ground-truth answer consists of $N$ letters (e.g., caterpillar). For the \textbf{dynamic task}, the first-attempt hint discloses the number of words in the answer and the number of letters in each word (e.g., \_ \_ \_ \_ \_ \_ \_ \_ \_ \_ \_). In the second attempt, based on the previous hint, we additionally reveal $\lceil N/8 \rceil$ randomly selected letters (e.g., \_ \_ \_ \_  r \_ i \_ \_ \_ \_). The third-attempt hint builds on the second attempt by further providing an additional $\lceil N/4 \rceil - \lceil N/8 \rceil$ randomly chosen letters (e.g., c \_ \_ \_ r \_ i \_ \_ \_ \_). For the \textbf{static task}, we directly adopt the third-attempt hint used in the dynamic task as the sole hint for the question. 

In addition to the symbolic representation, we also provide the hints in natural language to aid VLMs in better interpreting the partial information (e.g., The answer consists of 1 word. The 1st word has 11 letters. The 5th letter is `r'. The 7th letter is `i'.). Two illustrative examples for both the dynamic and static tasks are presented in Appendix~\ref{appendix:static_dynamic_eg}.

The choice of the divisors 8 and 4 serves two purposes: (1) to better mimic the hint progression in the original ``Guess the Build'' game, and (2) to balance the difficulty across different answer lengths, thereby enabling a more nuanced evaluation of model performance on the \ourdataset{}.

\subsection{Evaluation Metrics} 
\label{2.3: Evaluation metric}

To evaluate the responses generated by VLMs, we first follow the methodology proposed in MathVista~\cite{lu2023mathvista} by employing GPT-4o to extract the predicted guess from each response (Appendix~\ref{Appendix:ans_extractor}), and subsequently adopt accuracy as the evaluation metric. We observe that the median and mean answer lengths across all build sets are 8 and 8.15 letters respectively. Motivated by this observation, we categorize questions with answers of length less than or equal to 8 as \emph{short} answer questions, and those with longer answers as \emph{long} answer questions. In addition to reporting the overall accuracy, we also report the accuracy separately for the short and long subsets, aiming to investigate whether the length of the concept word is an impact factor.

Specifically, for the three-attempt dynamic tasks, if an VLM correctly answers a build-set question in either the first or second attempt, it is exempted from answering the question again in subsequent attempts, which are automatically marked as correct.

\subsection{Ethical Considerations} 
\label{section:ethical_considerations}

We take various steps to ensure the ethical compliance of \ourdataset{}. We first inspect the Minecraft terms of service\footnote{\url{https://www.minecraft.net/en-us/terms/r2}} for multiplayer servers as well as the Hypixel server's terms of service\footnote{\url{https://hypixel.net/terms}}, ensuring that players consent to be viewed by a larger audience when joining multiplayer servers and the academic use and anonymized resharing of in-game content is within intended use. To protect player privacy, we apply blurring to any visible player IDs for anonymization. We manually inspect all collected images and remove any that feature hateful or offensive builds.

\section{Experiment Settings}
\label{section:exp_settings}

\subsection{Models and Implementation} 
\label{3.1:models_and_implementation}

We evaluate six widely used VLMs on \ourdataset{}, including three API and three open models. The API models include GPT-4o~\cite{gpt4o}, GPT-4o-mini~\cite{gpt4o}, and Gemini-2.0-Flash~\cite{team2023gemini}. For the open models, we evaluate InternVL2.5-78B-MPO~\cite{wang2024enhancing}, Qwen2.5VL-72B~\cite{bai2025qwen2}, and MiniCPM-V2.6~\cite{yao2024minicpm}. 

To ensure the reproducibility, we standardize the decoding configurations across all VLMs by setting the temperature to 0.0 and top\_p to 1.0, or by disabling sampling via setting do\_sample to False. The specific prompts and hyperparameters used for each VLM are detailed in Appendix~\ref{appendix:exp_details}. All experiments are conducted using eight NVIDIA A100-SXM4-40GB GPUs.


\subsection{Reasoning Approaches} 
\label{3.2:reasoning_approach}

The models we tested employ chain-of-thought (CoT) reasoning by default, while we further explore several alternative reasoning strategies specifically for the strongest model GPT-4o. The results are reported in Table~\ref{tab:main_result}, and the corresponding instructions for each approach are provided in the Appendix~\ref{Appendix:reasoning_approaches}.

In the \textbf{w/o CoT} configuration, the prompt explicitly instructs the model to generate a direct answer without any intermediate reasoning steps. We explore this setup to investigate whether the intermediate reasoning steps are helpful in the task of creativity sensemaking.

In the \textbf{One-shot} setting, we provide a single demonstration. For \emph{static} tasks, this demonstration includes an image of a Minecraft build, a hint, and the corresponding answer. For \emph{dynamic} tasks, the demonstration consists of the current attempt along with all previous attempts, each accompanied by its associated image, hint, and answer.

The \textbf{Self-Consistency}~\cite{wang2022self} generates three independent responses and adopts the final prediction through majority voting. If all three responses differ, the last guess is selected as the final answer.

The \textbf{Self-Refine}~\cite{madaan2023self} approach prompts the model to evaluate its own response. If it deems the initial prediction correct, it outputs the answer directly. Otherwise, it reconsiders its response up to two additional times. If the model is still unable to affirm the correctness of any prediction, the last guess is adopted as the final answer.

Building on this, and tailored to the unique characteristics of our task, we further propose an extension of the Self-Refine method that incorporates \textbf{Image Retrieval}~\citep{zhu2024enhancing} to better support the model's self-evaluation. During self-evaluation, the model uses its current guess as a query to retrieve the top image result from Google Images\footnote{\url{https://www.google.com/imghp}.}. This retrieved image is then incorporated into the model’s reasoning to reassess the validity of its prediction.

\section{Results}
\label{4:results}

\begin{table*}[t!]
\resizebox{1\linewidth}{!}{
      \begin{tabular}{lrrarrrarrrarrrar@{}}
        \hline
        \toprule
        \multirow{3}{*}{\textbf{Model}} &
        \multicolumn{3}{c}{\textbf{Static}} & \phantom{} &
        \multicolumn{3}{c}{\textbf{Dynamic \#1}} & \phantom{} &
        \multicolumn{3}{c}{\textbf{Dynamic \#2}} & \phantom{} &
        \multicolumn{3}{c}{\textbf{Dynamic \#3}} \\
        
        \cmidrule{2-4} \cmidrule{6-8} \cmidrule{10-12} \cmidrule{14-16}
        
        \phantom{} & \small \textbf{Short} & \small \textbf{Long} &  \small \textbf{All} &  & \small \textbf{Short} & \small \textbf{Long} &  \small \textbf{All} &  & \small \textbf{Short} & \small \textbf{Long} &  \small \textbf{All} &  & \small \textbf{Short} & \small \textbf{Long} &  \small \textbf{All}  \\

        \midrule
        
        \multicolumn{16}{c}{\textbf{API Models}} \\
        
        \midrule

        GPT-4o &  \textbf{58.7} & \textbf{56.8} & \textbf{57.8} &   & \textbf{10.8} & \textbf{12.0} & \textbf{11.4} &  & \textbf{40.2} & \textbf{44.4} & \textbf{42.2} &  & \textbf{65.6} & \textbf{66.4} & \textbf{66.0} \\
        
        Gemini-2.0-Flash &  47.5 & 44.0 & 45.8 &   & 8.5 & 7.5 & 8.0 &  & 29.3 & 25.7 & 27.6 &  & 52.9 & 50.6 & 51.8 \\

         GPT-4o-mini &  40.9 & 32.4 & 36.8 &   & 7.3 & 7.1 & 7.2 &  & 29.0 & 23.7 & 26.4 &  & 49.8 & 37.8 & 44.0 \\

        \midrule
        
        \multicolumn{16}{c}{\textbf{Open Models}} \\
        
        \midrule

        InternVL2.5-MPO &  \textbf{25.5} & \textbf{18.7} & \textbf{22.2} &   & \textbf{3.1} & \textbf{2.1} & \textbf{2.6} &  & \textbf{12.0} & \textbf{7.9} & \textbf{10.0} &  & \textbf{23.9} & \textbf{14.1} & \textbf{19.2} \\

        Qwen2.5VL-72B & 15.4 & 8.3 & 12.0 &   & 2.3 & 0.8 & 1.6 &  & 8.5 & 6.2 & 7.4 &  & 18.1 & 9.5 & 14.0 \\

        MiniCPM-V2.6 & 12.4 & 6.2 & 9.4 &   & 1.9 & 1.2 & 1.6 &  & 4.6 & 3.3 & 4.0 &  & 10.8 & 5.8 & 8.4 \\

        \midrule

        \multicolumn{16}{c}{\textbf{Single-Modality Inputs (GPT-4o)}} \\

        \midrule

        Visual Input Only & \textbf{23.2} & \textbf{22.0} & \textbf{22.6} &    & \textbf{6.9} & \textbf{6.2} & \textbf{6.6} &  & \textbf{17.0} & \textbf{14.5} & \textbf{15.8} &  & \textbf{30.1} & 26.6 & \textbf{28.4} \\
        Hint Only &  17.0 & 18.3 & 17.6 & 
        & 0.4 & 1.7 & 1.0 &  & 7.7 & 12.0 & 
        9.8 &  & 20.8 & \textbf{28.2} & 24.4 \\

        \midrule
        
        \multicolumn{16}{c}{\textbf{Augmented GPT-4o}} \\
        
        \midrule
        
        w/o CoT & 59.1 & \textbf{58.1} & 58.6 &   & \textbf{11.6} & 12.0 & 11.8 &  & 42.1 & 44.4 & 43.2 &  & 68.0 & \textbf{67.6} & 67.8 \\

        One-shot & 59.8 & 53.1 & 56.6 &   & 10.4 & 11.2 & 10.8 &  & 42.5 & 42.7 & 42.6 &  & \textbf{70.7} & 64.7 & 67.8 \\

        Self-Consistency & 59.1 & 55.2 & 57.2 &   & 11.2 & \textbf{13.3} & \textbf{12.2} &  & 39.4 & 42.3 & 40.8 &  & \textbf{70.7} & 65.6 & 68.2 \\

        Self-Refine & \textbf{61.0} & \textbf{58.1} & \textbf{59.6} &   & 9.3 & 10.4 & 9.8 &  & \textbf{43.2} & \textbf{46.1} & \textbf{44.6} &  & 70.3 & \textbf{67.6} & \textbf{69.0} \\

        Image Retrieval & 58.3 & 53.9 & 56.2 &  & 9.3 & 12.0 & 10.6 &  & 42.1 & 42.3 & 42.2 &  & 67.6 & 65.6 & 66.6 \\
        
        \bottomrule
        \hline
      \end{tabular}
    }
    \caption{Evaluation results on \ourdataset{}. \textbf{Bold} values indicate the best performance within each category. \textbf{All} denotes the overall accuracy for each task. One model name is abbreviated due to space limits. InternVL2.5-MPO: InternVL2.5-78B-MPO. The results demonstrate the challenging nature of \ourdataset{} and reveal a clear performance disparity between open and API models.}
  \vspace{-2ex}
  \label{tab:main_result}
\end{table*}

Table~\ref{tab:main_result} presents the performance of six open and API models on \ourdataset{}, from which we draw several key observations.

\textbf{First, \ourdataset{} is highly challenging.} The best-performing model GPT-4o achieves only 57.8\% accuracy on static tasks and 66.0\% accuracy on the final attempt of dynamic tasks. This indicates that in at least 34.0\% of cases, even the strongest VLM wasn't able to decipher user creativity.

\textbf{Second, we observe a significant performance gap between open and API models.} GPT-4o outperforms the best open-source model by 160\% in static tasks and by 244\% in the third attempt of dynamic tasks. This disparity indicates that, although open models demonstrate competitive or even superior performance to API models on earlier benchmarks~\cite{bai2025qwen2, wang2024enhancing}, their performance degrades substantially on \ourdataset{}’s novel tasks. This suggests that previous benchmark success may stem from memorization of similar data and tasks, rather than a genuine and overall improvement of visual perception and reasoning. Consequently, open models exhibit limited creative and conceptual understanding.

\textbf{Third, API models benefit considerably from the iterative refinement in the dynamic setting.} In contrast, open models show minimal improvement, and in some cases, performance even declines. For the third attempt in the dynamic task, models are additionally provided with less complete Minecraft builds and imperfect hints from the previous two attempts. Under this setting, the average accuracy of API models increases by 15.24\% compared to their static task performance. However, open models exhibit little to no improvement, and the performance of InternVL2.5-78B-MPO even decreases by 13.51\%. This indicates that open models struggle to understand the incomplete builds in \ourdataset{}, further revealing their limited ability to generalize to unfamiliar tasks.

\textbf{Fourth, none of the reasoning strategies tested leads to substantial performance improvements.} Among them, Self-Refine proves to be most effective, while Image Retrieval performs poorly. Self-Refine increases accuracy by only 3.11\% on static tasks and 4.55\% on the third attempt of dynamic tasks. This outcome suggests that the challenges posed by \ourdataset{} cannot be adequately addressed through existing strategies to enhance VLMs. Interestingly, Image Retrieval causes a 2.77\% decrease in static task accuracy and only yields a 0.91\% increase in dynamic task attempt 3. Our in-depth case study reveals that some questions in \ourdataset{} have multiple valid answers, each corresponding to a different visual representation. When a model’s initial guess is correct, the retrieved image may reflect an alternative but still valid construction. However, due to differences between the original and retrieved images, the model may become confused and abandon its correct initial guess (Appendix~\ref{appendix:image_retrieval_fail}). This highlights a current limitation in the model’s ability to handle many-to-one mappings between images and textual answers, particularly in scenarios requiring creative visual perception. It also indicates that while most existing VLM enhancement methods on evaluated on common domains such as VQA and math, progress might not be generalizable to the wide spectrum of other VLM uses such as comprehending user creativity.

\textbf{Lastly, \ourdataset{} is fundamentally a cross-modal benchmark, where relying solely on text or image input is insufficient.} On static tasks, GPT-4o achieves only 22.6\% accuracy when given image-only input and 17.6\% accuracy with text-only input. These are both substantially lower than the 57.8\% accuracy achieved with multimodal input. A similar pattern is observed in dynamic tasks. This demonstrates that a single modality alone is inadequate for solving the tasks in \ourdataset{}, highlighting the necessity of effective multimodal integration for successful task completion.

\section{Analysis}

\subsection{Quantitative Evaluation of \ourdataset{} via AutoBencher Metrics}

\begin{wrapfigure}{r}{0.5\textwidth}
    \vspace{-26pt}
  \begin{center}
    \includegraphics[width=\linewidth]{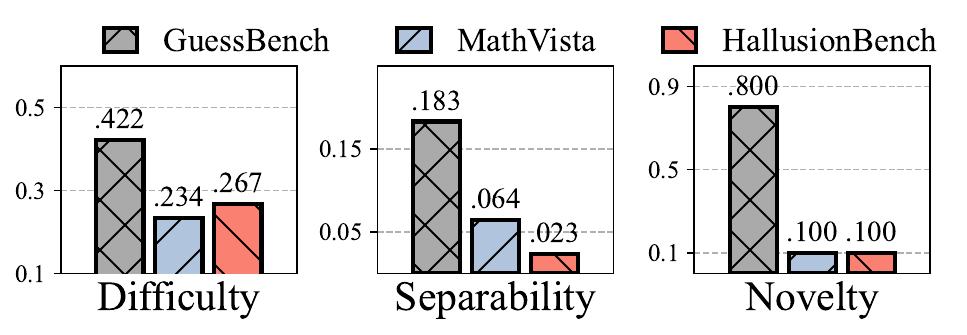}
  \end{center}
  \vspace*{-12pt}
  \caption{The AutoBencher scores of \ourdataset{}, MathVista, and HallusionBench, higher is better. \ourdataset{} demonstrates higher levels of difficulty, separability, and novelty compared to the other two benchmarks.}
  \vspace{-5pt}
  \label{fig: autobench_metrics}
\end{wrapfigure}

We adopt the evaluation metrics proposed by the AutoBencher~\cite{liautobencher} framework to quantitatively assess the quality of our dataset. Specifically, we compute three key metrics: Difficulty, Separability, and Novelty. The \textbf{Difficulty} of a benchmark is defined as the lowest error rate achieved by any model, reflecting the overall challenge posed by the task. \textbf{Separability} measures the degree to which models can be distinguished based on their performance, calculated as the mean absolute deviation of accuracies across models. \textbf{Novelty} quantifies how distinct the benchmark is from existing ones, computed as one minus the rank correlation between the accuracy vector on the current dataset and the most similar predicted accuracy vector derived from existing datasets. These metrics provide a quantitative measurement of dataset and task quality.

To compute these metrics, we evaluate five vision-language models: two API models (GPT-4o~\cite{gpt4o} and Gemini-2.0-Flash~\cite{team2023gemini}) and three open models (InternVL2.5-78B-MPO~\cite{wang2024enhancing}, Qwen2.5VL-72B~\cite{bai2025qwen2}, and MiniCPM-V2.6~\cite{yao2024minicpm}). We benchmark \ourdataset{} alongside MathVista~\cite{lu2023mathvista} and HallusionBench~\cite{guan2024hallusionbench}, with the results summarized in Figure~\ref{fig: autobench_metrics}. The detailed accuracy of all models on all involved benchmarks is presented in Appendix~\ref{appendix:detailed_accuracy}.

\paragraph{Difficulty.} \ourdataset{} achieves a difficulty score of 0.422, which is 80.34\% higher than MathVista and 58.05\% higher than HallusionBench. This significant increase indicates the heightened complexity and challenge posed by \ourdataset{}.

\paragraph{Separability.} \ourdataset{} obtains a separability score of 0.183, which is 2.86 times that of MathVista and 7.96 times that of HallusionBench. This suggests that \ourdataset{} is substantially more effective at differentiating the performance of various models in understanding creativity in the wild, revealing the strengths and weaknesses of models that were unclear with previous datasets.

\paragraph{Novelty.} We assess novelty by considering MMBench V1.1 Test (EN)~\cite{liu2024mmbench} and AI2D~\cite{kembhavi2016diagram} as prior datasets. Based on this metric, \ourdataset{} achieves higher novelty score than both MathVista and HallusionBench. Moreover, the model-specific accuracy scores on \ourdataset{} reveal a clear performance gap between open-source and proprietary models in addressing creative reasoning tasks. This further underscores the novelty and significance of our dataset and benchmark task in evaluating real-world multimodal understanding.

\subsection{Evaluating Multilingual Robustness in Creative Tasks}
\begin{figure}[t]
    \centering
    \includegraphics[width=\linewidth]{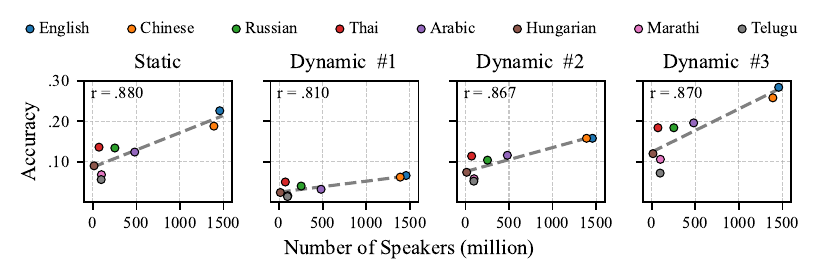}
    \vspace{-22pt}
    \caption{Performance of GPT-4o across eight languages with varying levels of resource availability. In the chart, r denotes the Pearson correlation coefficient. For low-resource languages, GPT-4o exhibits a noticeable decline in performance across four different settings. The performance shows a strong correlation with how widely each language is spoken.}
    \vspace{-13pt}
    \label{fig: multilingual}
\end{figure}
To investigate whether VLMs exhibit performance disparities across different linguistic environments, we conduct a multilingual evaluation based on the tasks categorized as Visual Input Only in Table~\ref{tab:main_result}. For each selected task, we provide identical prompts translated into various languages and translate the corresponding answers into the same target languages. GPT-4o is then tasked with reasoning under different language settings of the same problem.

To quantify language prevalence, we refer to the number of speakers (in millions) for each language as reported by Wikipedia, and visualize the relationship in Figure~\ref{fig: multilingual}. We observe that for high-resource languages such as English and Chinese, GPT-4o achieves significantly higher accuracy across all three attempts on both static and dynamic tasks, compared to languages with fewer speakers. Across all tasks, the Pearson correlation coefficient between the number of speakers and GPT-4o's accuracy is at least 0.810, indicating a strong positive correlation.

These findings suggest that the creative sensemaking capabilities of VLMs are substantially reduced in low-resource language settings, revealing a critical gap in current multilingual generalization and raising concerns about the equity of model performance when serving diverse language speakers.

\subsection{Impact of Concept Frequency}

\begin{wrapfigure}{r}{0.5\textwidth}
    \vspace{-40pt}
  \begin{center}
    \includegraphics[width=\linewidth]{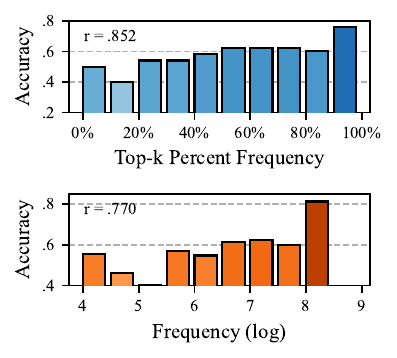}
  \end{center}
  \vspace{-18pt}
  \caption{GPT-4o performance across different levels of concept frequency. Bins containing fewer than three answers are omitted and shown as blank. Darker bin colors indicate higher accuracy values. In the chart, r denotes the Pearson correlation coefficient. Both plots reveal a consistent trend: GPT-4o's accuracy increases with concept frequency.}
  \vspace{-20pt}
  \label{fig: infigram}
\end{wrapfigure}

To investigate the impact of concept frequency on model performance, we adopt the Infini-gram~\cite{Liu2024InfiniGram} approach and use Dolma-v1.7 (2.6T tokens) as our corpus to obtain concept frequency statistics for each problem. We then evaluate the accuracy of GPT-4o under the static task setting. The accuracy for each bin represents the average accuracy of all samples within that range. As shown in Figure~\ref{fig: infigram}, we visualize the results using two methods: on the left, the x-axis represents the top-k percentage of answers sorted in ascending order of frequency; on the right, the x-axis directly reflects the raw word frequency in ascending order. We compute the Pearson correlation coefficient between accuracy and average word frequency across bins, yielding values of 0.852 and 0.770 respectively, indicating a strong positive correlation. These results suggest that when GPT-4o attempts to understand creativity in the wild, it is more likely to succeed with higher-frequency concepts, revealing limitations in its sensemaking capabilities for long-tail entities and concepts underrepresented in the training data.

\subsection{Effect of Multi-View Inputs}
\begin{figure}[t]
    \centering
    \includegraphics[width=\linewidth]{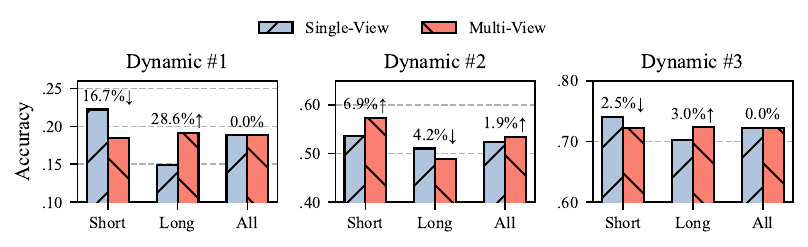}
    \vspace{-20pt}
    \caption{GPT-4o performance under Single-View and Multi-View settings. In three successive attempts of the Dynamic task, having Minecraft build images from different angles affects GPT-4o’s performance on both short-answer and long-answer questions. However, as the number of attempts increases, the impact gradually diminishes, while the overall performance remains largely consistent.}
    \vspace{-10pt}
    \label{fig: multiview}
\end{figure}

To explore whether providing VLMs with additional physical views of the same Minecraft build enhances their creative sensemaking, we evaluate GPT-4o under the Dynamic task setting. Specifically, we sample 101 sets of Minecraft build images, and for each attempt within the Dynamic task, we supplement the original input with two additional images captured from different viewpoints of the same build.

As shown in Figure~\ref{fig: multiview}, an intriguing pattern emerges: when analyzing short-answer and long-answer questions individually, additional viewpoints have some influence on GPT-4o's accuracy, but this influence decreases markedly with more attempts. In contrast, when considering overall accuracy across both question types, the impact of additional viewpoints becomes minimal or even negligible.

These findings suggest that merely offering more visual angles of a build does not substantially aid in identifying what the build represents, echoing VLMs' limitations in multi-view reasoning \citep{zhang2025do}. This further highlights the challenging nature of \ourdataset{}, as it requires VLMs to engage in deeper creative understanding, association, and reasoning rather than relying solely on expanded visual input.

\subsection{Impact of \ourdataset{} Fine-Tuning on Related Tasks}

\begin{wrapfigure}{r}{0.5\textwidth}
  \vspace{-22pt}
  \begin{center}
    \includegraphics[width=\linewidth]{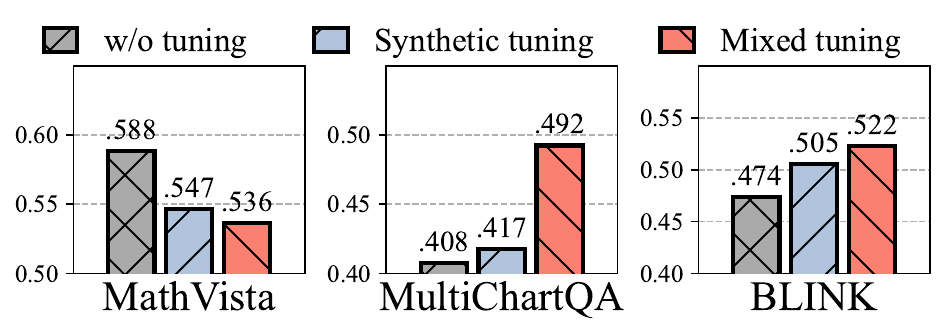}
  \end{center}
  \vspace{-15pt}
  \caption{The performance of various fine-tuning methods. Fine-tuning improves performance on both MultiChartQA and BLINK, whereas it reduces performance on MathVista.}
  \vspace{-7pt}
  \label{fig: ft_results}
\end{wrapfigure}

To investigate whether creativity sensemaking benefits VLMs on other tasks, we conduct a series of transfer learning experiments using the Qwen2-VL-7B model. We evaluate performance under three distinct settings: without tuning, synthetic tuning, and mixed tuning, across three benchmarks: MathVista~\cite{lu2023mathvista}, MultiChartQA~\cite{zhu2024multichartqa}, and BLINK~\cite{fu2024blink}. Detailed results are shown in Figure~\ref{fig: ft_results}.

For the synthetic tuning setting, we collect all instances correctly answered by GPT-4o on \ourdataset{} under the static task setting, yielding 289 data points. We then fine-tune Qwen2-VL-7B using these questions and GPT-4o response pairs. For the mixed tuning setting, we augment the synthetic tuning data with an additional 289 examples sampled from tasks correctly answered by GPT-4o in the target benchmarks, resulting in a total of 578 training examples. For evaluation, we exclude these training samples from each benchmark and randomly select 289 held-out examples per benchmark as the test set. All fine-tuning hyperparameters are kept consistent across conditions, as detailed in Appendix~\ref{appendix:fine-tune_details}.

Our findings show that synthetic tuning leads to noticeable performance improvements on MultiChartQA and BLINK, although performance on MathVista declines. Under mixed tuning, accuracy improves further on MultiChartQA and BLINK, with particularly substantial gains on MultiChartQA, while MathVista performance continues to decline.

Since \ourdataset{} is designed to emphasize perceptual and interpretive capabilities, fine-tuning on it significantly boosts performance on perception-driven tasks like BLINK and, to a lesser extent, MultiChartQA. In contrast, it appears less beneficial and potentially detrimental for tasks that demand precise logical or mathematical reasoning, such as those in MathVista. Although both synthetic and mixed tuning promote CoT reasoning in Qwen2-VL-7B, the resulting gains are largely confined to tasks requiring high-level visual understanding and flexible sensemaking. These results suggest that \ourdataset{} effectively enhances a model's interpretive competence, making it better suited for tasks where nuanced perception and contextual inference are critical.

\section{Related Work}

\paragraph{VLM Evaluation.} VLMs are increasingly popular and demonstrate great performance across a wide range of vision-language tasks. As such, evaluating their capabilities has emerged as a critical research problem. Existing benchmarks predominantly focus on objective tasks, including: (1) General Question Answering, such as MMBench~\cite{liu2024mmbench}, SEED-Bench~\cite{li2023seed}, and MMMU~\cite{yue2024mmmu}; (2) Optical Character Recognition (OCR), such as OCR-VQA~\cite{mishra2019ocr} and OCRBench~\cite{liu2024ocrbench}; (3) Graphic Understanding, including AI2D~\cite{kembhavi2016diagram}, CharXiv~\cite{wang2024charxiv}, and MultiChartQA~\cite{zhu2024multichartqa}; and (4) Mathematics, such as MathVista~\cite{lu2023mathvista}, MathVision~\cite{wang2024measuring}, and MathV360K~\cite{shi2024math}. However, current VLM benchmarks fall short in systematically exploring the \emph{subjective} capabilities of these models. In this work, we propose to evaluate VLMs through tasks that assess understanding and sensemaking creativity through \ourdataset{}, offering a lens into the models' performance on more subjective dimensions.

\paragraph{Creativity in Generative Models.} Prior work on the creativity of generative models has primarily focused on elite-level creativity, such as high-quality writing~\cite{chakrabarty2024art, tian2024large, lu2024ai, lee2023comparative, kim-etal-2025-representation, gomez-rodriguez-williams-2023-confederacy, chen-ding-2023-probing, bae-kim-2024-collective, marco-etal-2024-pron, chen-etal-2024-hollmwood}, scientific discovery~\cite{kang2022augmenting, hope2022scaling}, image synthesis~\cite{huang2023composer, shah2025does, kamb2024analytic, isajanyansocial, feng2024redefining, lu2024procreate}, problem solving~\cite{tian2023macgyver, anonymous2025repurposinginai, alavi2023large, g2024discovering, nair-etal-2024-creative}, code generation~\cite{lu-etal-2025-benchmarking, ch-etal-2024-retrieval}, combinational creativity~\cite{zhong2024let, peng2025probing, nagarajanmulti, favero2025compositional, li2024tp2o}, and embodied reasoning~\cite{park2024mr, white2025collaborating, dong-etal-2024-villageragent, jia-etal-2025-simulbench, chaturvedi-etal-2024-nebula, li2024auto, qin2024mp5, li2025optimus}. However, prior research has largely overlooked creativity at the level of the general public, which can be described as creativity in the wild. While such creativity may be imperfect, it is widespread and highly personalized. To address this gap, we propose \ourdataset{} as a proxy for quantifying the ability of VLMs to understand the creativity in the wild.

\section{Conclusion}

We present \ourdataset{}, a novel benchmark designed to evaluate VLMs' capacity for creativity sensemaking in the wild. By leveraging gameplay data from the Minecraft minigame ``Guess the Build'', our dataset captures the pervasive, noisy, and pluralistic nature of real-world human creativity, offering a unique and challenging testbed for VLMs acting as guessers. Through comprehensive experiments, we demonstrate that state-of-the-art models such as GPT-4o struggle with dynamic reasoning and decoding linguistically/culturally underrepresented contexts. Our findings underscore the limitations of existing VLMs in modeling diverse, imperfect forms of creativity and highlight the importance of training on realistic, user-generated data. \ourdataset{} not only reveals nuanced gaps in model performance but also shows potential for enhancing VLM capabilities through fine-tuning on reasoning traces for creativity sensemaking. We envision \ourdataset{} as a valuable resource in aligning VLMs more closely with the complex and inclusive landscape of human creative expression.

\bibliography{neurips_2025}

\begin{thebibliography}{88}
\providecommand{\natexlab}[1]{#1}
\providecommand{\url}[1]{\texttt{#1}}
\expandafter\ifx\csname urlstyle\endcsname\relax
  \providecommand{\doi}[1]{doi: #1}\else
  \providecommand{\doi}{doi: \begingroup \urlstyle{rm}\Url}\fi

\bibitem[Alavi~Naeini et~al.(2023)Alavi~Naeini, Saqur, Saeidi, Giorgi, and Taati]{alavi2023large}
Saeid Alavi~Naeini, Raeid Saqur, Mozhgan Saeidi, John Giorgi, and Babak Taati.
\newblock Large language models are fixated by red herrings: Exploring creative problem solving and einstellung effect using the only connect wall dataset.
\newblock \emph{Advances in Neural Information Processing Systems}, 36:\penalty0 5631--5652, 2023.

\bibitem[Anonymous(2025)]{anonymous2025repurposinginai}
Anonymous.
\newblock Repurposing in ai: A distinct approach or an extension of creative problem solving?
\newblock In \emph{ICLR Blogposts 2025}, 2025.
\newblock URL \url{https://d2jud02ci9yv69.cloudfront.net/2025-04-28-repurposing-85/blog/repurposing/}.
\newblock https://d2jud02ci9yv69.cloudfront.net/2025-04-28-repurposing-85/blog/repurposing/.

\bibitem[Bae and Kim(2024)]{bae-kim-2024-collective}
Minwook Bae and Hyounghun Kim.
\newblock Collective critics for creative story generation.
\newblock In Yaser Al-Onaizan, Mohit Bansal, and Yun-Nung Chen, editors, \emph{Proceedings of the 2024 Conference on Empirical Methods in Natural Language Processing}, pages 18784--18819, Miami, Florida, USA, November 2024. Association for Computational Linguistics.
\newblock \doi{10.18653/v1/2024.emnlp-main.1046}.
\newblock URL \url{https://aclanthology.org/2024.emnlp-main.1046/}.

\bibitem[Bai et~al.(2025)Bai, Chen, Liu, Wang, Ge, Song, Dang, Wang, Wang, Tang, et~al.]{bai2025qwen2}
Shuai Bai, Keqin Chen, Xuejing Liu, Jialin Wang, Wenbin Ge, Sibo Song, Kai Dang, Peng Wang, Shijie Wang, Jun Tang, et~al.
\newblock Qwen2. 5-vl technical report.
\newblock \emph{arXiv preprint arXiv:2502.13923}, 2025.

\bibitem[Barton(2013)]{barton2013arts}
Georgina Barton.
\newblock The arts and literacy: What does it mean to be arts literate?
\newblock \emph{International Journal of Education \& the Arts}, 14\penalty0 (18), 2013.

\bibitem[Chakrabarty et~al.(2024)Chakrabarty, Laban, Agarwal, Muresan, and Wu]{chakrabarty2024art}
Tuhin Chakrabarty, Philippe Laban, Divyansh Agarwal, Smaranda Muresan, and Chien-Sheng Wu.
\newblock Art or artifice? large language models and the false promise of creativity.
\newblock In \emph{Proceedings of the 2024 CHI Conference on Human Factors in Computing Systems}, pages 1--34, 2024.

\bibitem[Chaturvedi et~al.(2024)Chaturvedi, Thompson, and Asher]{chaturvedi-etal-2024-nebula}
Akshay Chaturvedi, Kate Thompson, and Nicholas Asher.
\newblock Nebula: A discourse aware {M}inecraft builder.
\newblock In Yaser Al-Onaizan, Mohit Bansal, and Yun-Nung Chen, editors, \emph{Findings of the Association for Computational Linguistics: EMNLP 2024}, pages 6431--6443, Miami, Florida, USA, November 2024. Association for Computational Linguistics.
\newblock \doi{10.18653/v1/2024.findings-emnlp.374}.
\newblock URL \url{https://aclanthology.org/2024.findings-emnlp.374/}.

\bibitem[Chen and Ding(2023)]{chen-ding-2023-probing}
Honghua Chen and Nai Ding.
\newblock Probing the {\textquotedblleft}creativity{\textquotedblright} of large language models: Can models produce divergent semantic association?
\newblock In Houda Bouamor, Juan Pino, and Kalika Bali, editors, \emph{Findings of the Association for Computational Linguistics: EMNLP 2023}, pages 12881--12888, Singapore, December 2023. Association for Computational Linguistics.
\newblock \doi{10.18653/v1/2023.findings-emnlp.858}.
\newblock URL \url{https://aclanthology.org/2023.findings-emnlp.858/}.

\bibitem[Chen et~al.(2024)Chen, Zhu, Yang, Shi, Xi, Zhang, Wang, Pu, Feng, Yang, and Zhang]{chen-etal-2024-hollmwood}
Jing Chen, Xinyu Zhu, Cheng Yang, Chufan Shi, Yadong Xi, Yuxiang Zhang, Junjie Wang, Jiashu Pu, Tian Feng, Yujiu Yang, and Rongsheng Zhang.
\newblock {H}o{LLM}wood: Unleashing the creativity of large language models in screenwriting via role playing.
\newblock In Yaser Al-Onaizan, Mohit Bansal, and Yun-Nung Chen, editors, \emph{Findings of the Association for Computational Linguistics: EMNLP 2024}, pages 8075--8121, Miami, Florida, USA, November 2024. Association for Computational Linguistics.
\newblock \doi{10.18653/v1/2024.findings-emnlp.474}.
\newblock URL \url{https://aclanthology.org/2024.findings-emnlp.474/}.

\bibitem[Cherep et~al.(2024)Cherep, Singh, and Shand]{cherep2024creative}
Manuel Cherep, Nikhil Singh, and Jessica Shand.
\newblock Creative text-to-audio generation via synthesizer programming.
\newblock In \emph{Proceedings of the 41st International Conference on Machine Learning}, pages 8270--8285, 2024.

\bibitem[Daniel~Han and team(2023)]{unsloth}
Michael~Han Daniel~Han and Unsloth team.
\newblock Unsloth, 2023.
\newblock URL \url{http://github.com/unslothai/unsloth}.

\bibitem[Dong et~al.(2024)Dong, Zhu, Pan, Zhu, and Yang]{dong-etal-2024-villageragent}
Yubo Dong, Xukun Zhu, Zhengzhe Pan, Linchao Zhu, and Yi~Yang.
\newblock {V}illager{A}gent: A graph-based multi-agent framework for coordinating complex task dependencies in {M}inecraft.
\newblock In Lun-Wei Ku, Andre Martins, and Vivek Srikumar, editors, \emph{Findings of the Association for Computational Linguistics: ACL 2024}, pages 16290--16314, Bangkok, Thailand, August 2024. Association for Computational Linguistics.
\newblock \doi{10.18653/v1/2024.findings-acl.964}.
\newblock URL \url{https://aclanthology.org/2024.findings-acl.964/}.

\bibitem[Fang et~al.(2025)Fang, Chen, Lan, Ding, Liang, Zhao, Wen, Zhang, Zhang, Duan, et~al.]{fang2025creation}
Xinyu Fang, Zhijian Chen, Kai Lan, Shengyuan Ding, Yingji Liang, Xiangyu Zhao, Farong Wen, Zicheng Zhang, Guofeng Zhang, Haodong Duan, et~al.
\newblock Creation-mmbench: Assessing context-aware creative intelligence in mllm.
\newblock \emph{arXiv preprint arXiv:2503.14478}, 2025.

\bibitem[Favero et~al.()Favero, Sclocchi, Cagnetta, Frossard, and Wyart]{favero2025compositional}
Alessandro Favero, Antonio Sclocchi, Francesco Cagnetta, Pascal Frossard, and Matthieu Wyart.
\newblock How compositional generalization and creativity improve as diffusion models are trained.
\newblock In \emph{ICLR 2025 Workshop on Deep Generative Model in Machine Learning: Theory, Principle and Efficacy}.

\bibitem[Feng et~al.(2024{\natexlab{a}})Feng, Xie, Yang, Wang, and Geng]{feng2024redefining}
Fu~Feng, Yucheng Xie, Xu~Yang, Jing Wang, and Xin Geng.
\newblock Redefining< creative> in dictionary: Towards an enhanced semantic understanding of creative generation.
\newblock \emph{arXiv preprint arXiv:2410.24160}, 2024{\natexlab{a}}.

\bibitem[Feng et~al.(2024{\natexlab{b}})Feng, Weng, Wang, Yuan, Bao, Luo, Chen, and Guo]{feng2024ccedit}
Ruoyu Feng, Wenming Weng, Yanhui Wang, Yuhui Yuan, Jianmin Bao, Chong Luo, Zhibo Chen, and Baining Guo.
\newblock Ccedit: Creative and controllable video editing via diffusion models.
\newblock In \emph{Proceedings of the IEEE/CVF Conference on Computer Vision and Pattern Recognition}, pages 6712--6722, 2024{\natexlab{b}}.

\bibitem[Feng et~al.(2024{\natexlab{c}})Feng, Sorensen, Liu, Fisher, Park, Choi, and Tsvetkov]{feng2024modular}
Shangbin Feng, Taylor Sorensen, Yuhan Liu, Jillian Fisher, Chan~Young Park, Yejin Choi, and Yulia Tsvetkov.
\newblock Modular pluralism: Pluralistic alignment via multi-llm collaboration.
\newblock In \emph{Proceedings of the 2024 Conference on Empirical Methods in Natural Language Processing}, pages 4151--4171, 2024{\natexlab{c}}.

\bibitem[Fu et~al.(2024)Fu, Hu, Li, Feng, Wang, Lin, Roth, Smith, Ma, and Krishna]{fu2024blink}
Xingyu Fu, Yushi Hu, Bangzheng Li, Yu~Feng, Haoyu Wang, Xudong Lin, Dan Roth, Noah~A Smith, Wei-Chiu Ma, and Ranjay Krishna.
\newblock Blink: Multimodal large language models can see but not perceive.
\newblock In \emph{European Conference on Computer Vision}, pages 148--166. Springer, 2024.

\bibitem[G~Leon et~al.(2024)G~Leon, Riccio, Subramanian, Wurman, and Stone]{g2024discovering}
Borja G~Leon, Francesco Riccio, Kaushik Subramanian, Peter Wurman, and Peter Stone.
\newblock Discovering creative behaviors through duplex: Diverse universal features for policy exploration.
\newblock \emph{Advances in Neural Information Processing Systems}, 37:\penalty0 49625--49648, 2024.

\bibitem[Gokaslan et~al.(2024)Gokaslan, Cooper, Collins, Seguin, Jacobson, Patel, Frankle, Stephenson, and Kuleshov]{gokaslan2024commoncanvas}
Aaron Gokaslan, A~Feder Cooper, Jasmine Collins, Landan Seguin, Austin Jacobson, Mihir Patel, Jonathan Frankle, Cory Stephenson, and Volodymyr Kuleshov.
\newblock Commoncanvas: Open diffusion models trained on creative-commons images.
\newblock In \emph{Proceedings of the IEEE/CVF Conference on Computer Vision and Pattern Recognition}, pages 8250--8260, 2024.

\bibitem[G{\'o}mez-Rodr{\'i}guez and Williams(2023)]{gomez-rodriguez-williams-2023-confederacy}
Carlos G{\'o}mez-Rodr{\'i}guez and Paul Williams.
\newblock A confederacy of models: a comprehensive evaluation of {LLM}s on creative writing.
\newblock In Houda Bouamor, Juan Pino, and Kalika Bali, editors, \emph{Findings of the Association for Computational Linguistics: EMNLP 2023}, pages 14504--14528, Singapore, December 2023. Association for Computational Linguistics.
\newblock \doi{10.18653/v1/2023.findings-emnlp.966}.
\newblock URL \url{https://aclanthology.org/2023.findings-emnlp.966/}.

\bibitem[Guan et~al.(2024)Guan, Liu, Wu, Xian, Li, Liu, Wang, Chen, Huang, Yacoob, et~al.]{guan2024hallusionbench}
Tianrui Guan, Fuxiao Liu, Xiyang Wu, Ruiqi Xian, Zongxia Li, Xiaoyu Liu, Xijun Wang, Lichang Chen, Furong Huang, Yaser Yacoob, et~al.
\newblock Hallusionbench: an advanced diagnostic suite for entangled language hallucination and visual illusion in large vision-language models.
\newblock In \emph{Proceedings of the IEEE/CVF Conference on Computer Vision and Pattern Recognition}, pages 14375--14385, 2024.

\bibitem[Han et~al.(2024)Han, Obieke, Zhao, and Jiang]{han2024using}
Ji~Han, Chijioke~C Obieke, Haosong Zhao, and Pingfei Jiang.
\newblock Using ai to generate short videos as stimuli for supporting design creativity.
\newblock In \emph{DS 136: Proceedings of the Asia Design and Innovation Conference (ADIC) 2024}. The Design Society, 2024.

\bibitem[Han et~al.(2025)Han, Kwon, Lee, Kim, and Choi]{han2025enhancing}
Jiyeon Han, Dahee Kwon, Gayoung Lee, Junho Kim, and Jaesik Choi.
\newblock Enhancing creative generation on stable diffusion-based models.
\newblock \emph{arXiv preprint arXiv:2503.23538}, 2025.

\bibitem[Hope et~al.(2022)Hope, Tamari, Hershcovich, Kang, Chan, Kittur, and Shahaf]{hope2022scaling}
Tom Hope, Ronen Tamari, Daniel Hershcovich, Hyeonsu~B Kang, Joel Chan, Aniket Kittur, and Dafna Shahaf.
\newblock Scaling creative inspiration with fine-grained functional aspects of ideas.
\newblock In \emph{Proceedings of the 2022 CHI Conference on Human Factors in Computing Systems}, pages 1--15, 2022.

\bibitem[Huang et~al.(2023)Huang, Chen, Liu, Shen, Zhao, and Zhou]{huang2023composer}
Lianghua Huang, Di~Chen, Yu~Liu, Yujun Shen, Deli Zhao, and Jingren Zhou.
\newblock Composer: creative and controllable image synthesis with composable conditions.
\newblock In \emph{Proceedings of the 40th International Conference on Machine Learning}, pages 13753--13773, 2023.

\bibitem[Isajanyan et~al.()Isajanyan, Shatveryan, Kocharian, Wang, and Shi]{isajanyansocial}
Arman Isajanyan, Artur Shatveryan, David Kocharian, Zhangyang Wang, and Humphrey Shi.
\newblock Social reward: Evaluating and enhancing generative ai through million-user feedback from an online creative community.
\newblock In \emph{The Twelfth International Conference on Learning Representations}.

\bibitem[Jia et~al.(2025)Jia, Yue, Zheng, Huang, and Lin]{jia-etal-2025-simulbench}
Qi~Jia, Xiang Yue, Tuney Zheng, Jie Huang, and Bill~Yuchen Lin.
\newblock {S}imul{B}ench: Evaluating language models with creative simulation tasks.
\newblock In Luis Chiruzzo, Alan Ritter, and Lu~Wang, editors, \emph{Findings of the Association for Computational Linguistics: NAACL 2025}, pages 8118--8131, Albuquerque, New Mexico, April 2025. Association for Computational Linguistics.
\newblock ISBN 979-8-89176-195-7.
\newblock URL \url{https://aclanthology.org/2025.findings-naacl.453/}.

\bibitem[Jiang et~al.(2025)Jiang, Zhang, Weller, Weir, Van~Durme, and Khashabi]{jiang2025self}
Dongwei Jiang, Jingyu Zhang, Orion Weller, Nathaniel Weir, Benjamin Van~Durme, and Daniel Khashabi.
\newblock Self-[in] correct: Llms struggle with discriminating self-generated responses.
\newblock In \emph{Proceedings of the AAAI Conference on Artificial Intelligence}, volume~39, pages 24266--24275, 2025.

\bibitem[Jiang et~al.(2024)Jiang, Yue, Luo, Ding, and Lu]{jiang2024reinforcement}
Haobin Jiang, Junpeng Yue, Hao Luo, Ziluo Ding, and Zongqing Lu.
\newblock Reinforcement learning friendly vision-language model for minecraft.
\newblock In \emph{European Conference on Computer Vision}, pages 1--17. Springer, 2024.

\bibitem[Kamb and Ganguli(2024)]{kamb2024analytic}
Mason Kamb and Surya Ganguli.
\newblock An analytic theory of creativity in convolutional diffusion models.
\newblock \emph{arXiv preprint arXiv:2412.20292}, 2024.

\bibitem[Kang et~al.(2022)Kang, Qian, Hope, Shahaf, Chan, and Kittur]{kang2022augmenting}
Hyeonsu~B Kang, Xin Qian, Tom Hope, Dafna Shahaf, Joel Chan, and Aniket Kittur.
\newblock Augmenting scientific creativity with an analogical search engine.
\newblock \emph{ACM Transactions on Computer-Human Interaction}, 29\penalty0 (6):\penalty0 1--36, 2022.

\bibitem[Kembhavi et~al.(2016)Kembhavi, Salvato, Kolve, Seo, Hajishirzi, and Farhadi]{kembhavi2016diagram}
Aniruddha Kembhavi, Mike Salvato, Eric Kolve, Minjoon Seo, Hannaneh Hajishirzi, and Ali Farhadi.
\newblock A diagram is worth a dozen images.
\newblock In \emph{Computer Vision--ECCV 2016: 14th European Conference, Amsterdam, The Netherlands, October 11--14, 2016, Proceedings, Part IV 14}, pages 235--251. Springer, 2016.

\bibitem[Kim et~al.(2025{\natexlab{a}})Kim, Sato, White, Ho, Lee, Hwang, and Mutlu]{kim2025bridging}
Callie~Y Kim, Arissa~J Sato, Nathan~Thomas White, Hui-Ru Ho, Christine~P Lee, Yuna Hwang, and Bilge Mutlu.
\newblock Bridging generations using ai-supported co-creative activities.
\newblock In \emph{Proceedings of the 2025 CHI Conference on Human Factors in Computing Systems}, pages 1--15, 2025{\natexlab{a}}.

\bibitem[Kim et~al.(2025{\natexlab{b}})Kim, Jo, On, and Lee]{kim-etal-2025-representation}
Deokgi Kim, Joonyoung Jo, Byung-Won On, and Ingyu Lee.
\newblock Representation-to-creativity ({R}2{C}): Automated holistic scoring model for essay creativity.
\newblock In Luis Chiruzzo, Alan Ritter, and Lu~Wang, editors, \emph{Findings of the Association for Computational Linguistics: NAACL 2025}, pages 5257--5275, Albuquerque, New Mexico, April 2025{\natexlab{b}}. Association for Computational Linguistics.
\newblock ISBN 979-8-89176-195-7.
\newblock URL \url{https://aclanthology.org/2025.findings-naacl.292/}.

\bibitem[Kranti et~al.(2024)Kranti, Hakimov, and Schlangen]{ch-etal-2024-retrieval}
Chalamalasetti Kranti, Sherzod Hakimov, and David Schlangen.
\newblock Retrieval-augmented code generation for situated action generation: A case study on {M}inecraft.
\newblock In Yaser Al-Onaizan, Mohit Bansal, and Yun-Nung Chen, editors, \emph{Findings of the Association for Computational Linguistics: EMNLP 2024}, pages 11159--11170, Miami, Florida, USA, November 2024. Association for Computational Linguistics.
\newblock \doi{10.18653/v1/2024.findings-emnlp.652}.
\newblock URL \url{https://aclanthology.org/2024.findings-emnlp.652/}.

\bibitem[Lee et~al.(2023)Lee, Kim, On, and Lee]{lee2023comparative}
Youbin Lee, Deokgi Kim, Byung-Won On, and Ingyu Lee.
\newblock A comparative analysis of the effectiveness of rare tokens on creative expression using rambert.
\newblock In \emph{Findings of the Association for Computational Linguistics: ACL 2023}, pages 10063--10077, 2023.

\bibitem[Li et~al.(2023)Li, Wang, Wang, Ge, Ge, and Shan]{li2023seed}
Bohao Li, Rui Wang, Guangzhi Wang, Yuying Ge, Yixiao Ge, and Ying Shan.
\newblock Seed-bench: Benchmarking multimodal llms with generative comprehension.
\newblock \emph{arXiv preprint arXiv:2307.16125}, 2023.

\bibitem[Li et~al.(2024{\natexlab{a}})Li, Yang, Wang, Zhu, Zhou, Qiao, Wang, Li, Lu, and Dai]{li2024auto}
Hao Li, Xue Yang, Zhaokai Wang, Xizhou Zhu, Jie Zhou, Yu~Qiao, Xiaogang Wang, Hongsheng Li, Lewei Lu, and Jifeng Dai.
\newblock Auto mc-reward: Automated dense reward design with large language models for minecraft.
\newblock In \emph{Proceedings of the IEEE/CVF Conference on Computer Vision and Pattern Recognition}, pages 16426--16435, 2024{\natexlab{a}}.

\bibitem[Li et~al.(2024{\natexlab{b}})Li, Zhang, and Yang]{li2024tp2o}
Jun Li, Zedong Zhang, and Jian Yang.
\newblock Tp2o: Creative text pair-to-object generation using balance swap-sampling.
\newblock In \emph{European Conference on Computer Vision}, pages 92--111. Springer, 2024{\natexlab{b}}.

\bibitem[Li et~al.()Li, Kaiyom, Liu, Mai, Liang, and Hashimoto]{liautobencher}
Xiang~Lisa Li, Farzaan Kaiyom, Evan~Zheran Liu, Yifan Mai, Percy Liang, and Tatsunori Hashimoto.
\newblock Autobencher: Towards declarative benchmark construction.
\newblock In \emph{The Thirteenth International Conference on Learning Representations}.

\bibitem[Li et~al.(2025)Li, Xie, Shao, Chen, Jiang, and Nie]{li2025optimus}
Zaijing Li, Yuquan Xie, Rui Shao, Gongwei Chen, Dongmei Jiang, and Liqiang Nie.
\newblock Optimus-2: Multimodal minecraft agent with goal-observation-action conditioned policy.
\newblock \emph{arXiv preprint arXiv:2502.19902}, 2025.

\bibitem[Lifshitz et~al.(2023)Lifshitz, Paster, Chan, Ba, and McIlraith]{lifshitz2023steve}
Shalev Lifshitz, Keiran Paster, Harris Chan, Jimmy Ba, and Sheila McIlraith.
\newblock Steve-1: A generative model for text-to-behavior in minecraft.
\newblock \emph{Advances in Neural Information Processing Systems}, 36:\penalty0 69900--69929, 2023.

\bibitem[Liu et~al.(2024{\natexlab{a}})Liu, Min, Zettlemoyer, Choi, and Hajishirzi]{Liu2024InfiniGram}
Jiacheng Liu, Sewon Min, Luke Zettlemoyer, Yejin Choi, and Hannaneh Hajishirzi.
\newblock Infini-gram: Scaling unbounded n-gram language models to a trillion tokens.
\newblock In \emph{First Conference on Language Modeling}, 2024{\natexlab{a}}.

\bibitem[Liu et~al.(2024{\natexlab{b}})Liu, Duan, Zhang, Li, Zhang, Zhao, Yuan, Wang, He, Liu, et~al.]{liu2024mmbench}
Yuan Liu, Haodong Duan, Yuanhan Zhang, Bo~Li, Songyang Zhang, Wangbo Zhao, Yike Yuan, Jiaqi Wang, Conghui He, Ziwei Liu, et~al.
\newblock Mmbench: Is your multi-modal model an all-around player?
\newblock In \emph{European conference on computer vision}, pages 216--233. Springer, 2024{\natexlab{b}}.

\bibitem[Liu et~al.(2024{\natexlab{c}})Liu, Li, Huang, Yang, Yu, Li, Yin, Liu, Jin, and Bai]{liu2024ocrbench}
Yuliang Liu, Zhang Li, Mingxin Huang, Biao Yang, Wenwen Yu, Chunyuan Li, Xu-Cheng Yin, Cheng-Lin Liu, Lianwen Jin, and Xiang Bai.
\newblock Ocrbench: on the hidden mystery of ocr in large multimodal models.
\newblock \emph{Science China Information Sciences}, 67\penalty0 (12):\penalty0 220102, 2024{\natexlab{c}}.

\bibitem[Lu et~al.(2024{\natexlab{a}})Lu, Teehan, and Ren]{lu2024procreate}
Jack Lu, Ryan Teehan, and Mengye Ren.
\newblock Procreate, don’t reproduce! propulsive energy diffusion for creative generation.
\newblock In \emph{European Conference on Computer Vision}, pages 397--414. Springer, 2024{\natexlab{a}}.

\bibitem[Lu et~al.(2024{\natexlab{b}})Lu, Bansal, Xia, Liu, Li, Hajishirzi, Cheng, Chang, Galley, and Gao]{lu2023mathvista}
Pan Lu, Hritik Bansal, Tony Xia, Jiacheng Liu, Chunyuan Li, Hannaneh Hajishirzi, Hao Cheng, Kai{-}Wei Chang, Michel Galley, and Jianfeng Gao.
\newblock Mathvista: Evaluating mathematical reasoning of foundation models in visual contexts.
\newblock In \emph{The Twelfth International Conference on Learning Representations, {ICLR} 2024, Vienna, Austria, May 7-11, 2024}, 2024{\natexlab{b}}.

\bibitem[Lu et~al.(2024{\natexlab{c}})Lu, Sclar, Hallinan, Mireshghallah, Liu, Han, Ettinger, Jiang, Chandu, Dziri, et~al.]{lu2024ai}
Ximing Lu, Melanie Sclar, Skyler Hallinan, Niloofar Mireshghallah, Jiacheng Liu, Seungju Han, Allyson Ettinger, Liwei Jiang, Khyathi Chandu, Nouha Dziri, et~al.
\newblock Ai as humanity's salieri: Quantifying linguistic creativity of language models via systematic attribution of machine text against web text.
\newblock \emph{arXiv preprint arXiv:2410.04265}, 2024{\natexlab{c}}.

\bibitem[Lu et~al.(2025)Lu, Wang, Li, Jiang, Khudanpur, Jiang, and Khashabi]{lu-etal-2025-benchmarking}
Yining Lu, Dixuan Wang, Tianjian Li, Dongwei Jiang, Sanjeev Khudanpur, Meng Jiang, and Daniel Khashabi.
\newblock Benchmarking language model creativity: A case study on code generation.
\newblock In Luis Chiruzzo, Alan Ritter, and Lu~Wang, editors, \emph{Proceedings of the 2025 Conference of the Nations of the Americas Chapter of the Association for Computational Linguistics: Human Language Technologies (Volume 1: Long Papers)}, pages 2776--2794, Albuquerque, New Mexico, April 2025. Association for Computational Linguistics.
\newblock ISBN 979-8-89176-189-6.
\newblock URL \url{https://aclanthology.org/2025.naacl-long.141/}.

\bibitem[Lv et~al.(2024)Lv, Gong, Liang, Pang, and Zhang]{lv-etal-2024-subjective}
Fangrui Lv, Kaixiong Gong, Jian Liang, Xinyu Pang, and Changshui Zhang.
\newblock Subjective topic meets {LLM}s: Unleashing comprehensive, reflective and creative thinking through the negation of negation.
\newblock In Yaser Al-Onaizan, Mohit Bansal, and Yun-Nung Chen, editors, \emph{Proceedings of the 2024 Conference on Empirical Methods in Natural Language Processing}, pages 12318--12341, Miami, Florida, USA, November 2024. Association for Computational Linguistics.
\newblock \doi{10.18653/v1/2024.emnlp-main.686}.
\newblock URL \url{https://aclanthology.org/2024.emnlp-main.686/}.

\bibitem[Madaan et~al.(2023)Madaan, Tandon, Gupta, Hallinan, Gao, Wiegreffe, Alon, Dziri, Prabhumoye, Yang, et~al.]{madaan2023self}
Aman Madaan, Niket Tandon, Prakhar Gupta, Skyler Hallinan, Luyu Gao, Sarah Wiegreffe, Uri Alon, Nouha Dziri, Shrimai Prabhumoye, Yiming Yang, et~al.
\newblock Self-refine: Iterative refinement with self-feedback.
\newblock \emph{Advances in Neural Information Processing Systems}, 36:\penalty0 46534--46594, 2023.

\bibitem[Marco et~al.(2024)Marco, Gonzalo, Mateo-Girona, and Santos]{marco-etal-2024-pron}
Guillermo Marco, Julio Gonzalo, M.Teresa Mateo-Girona, and Ram{\'o}n Del~Castillo Santos.
\newblock Pron vs prompt: Can large language models already challenge a world-class fiction author at creative text writing?
\newblock In Yaser Al-Onaizan, Mohit Bansal, and Yun-Nung Chen, editors, \emph{Proceedings of the 2024 Conference on Empirical Methods in Natural Language Processing}, pages 19654--19670, Miami, Florida, USA, November 2024. Association for Computational Linguistics.
\newblock \doi{10.18653/v1/2024.emnlp-main.1096}.
\newblock URL \url{https://aclanthology.org/2024.emnlp-main.1096/}.

\bibitem[Miller et~al.(2024)Miller, Dupont, and Wang]{miller2024enhanced}
Elijah Miller, Thomas Dupont, and Mingming Wang.
\newblock Enhanced creativity and ideation through stable video synthesis.
\newblock \emph{arXiv preprint arXiv:2405.13357}, 2024.

\bibitem[Minh et~al.()Minh, Baker, Neo, Roush, Kirsch, and Shwartz-Ziv]{minhturning}
Nguyen~Nhat Minh, Andrew Baker, Clement Neo, Allen~G Roush, Andreas Kirsch, and Ravid Shwartz-Ziv.
\newblock Turning up the heat: Min-p sampling for creative and coherent llm outputs.
\newblock In \emph{The Thirteenth International Conference on Learning Representations}.

\bibitem[Mishra et~al.(2019)Mishra, Shekhar, Singh, and Chakraborty]{mishra2019ocr}
Anand Mishra, Shashank Shekhar, Ajeet~Kumar Singh, and Anirban Chakraborty.
\newblock Ocr-vqa: Visual question answering by reading text in images.
\newblock In \emph{2019 international conference on document analysis and recognition (ICDAR)}, pages 947--952. IEEE, 2019.

\bibitem[Nagarajan et~al.()Nagarajan, Wu, Ding, and Raghunathan]{nagarajanmulti}
Vaishnavh Nagarajan, Chen~Henry Wu, Charles Ding, and Aditi Raghunathan.
\newblock Multi-token prediction boosts creativity in algorithmic tasks.
\newblock In \emph{Workshop on Spurious Correlation and Shortcut Learning: Foundations and Solutions}.

\bibitem[Nair et~al.(2024)Nair, Gizzi, and Sinapov]{nair-etal-2024-creative}
Lakshmi Nair, Evana Gizzi, and Jivko Sinapov.
\newblock Creative problem solving in large language and vision models - what would it take?
\newblock In Yaser Al-Onaizan, Mohit Bansal, and Yun-Nung Chen, editors, \emph{Findings of the Association for Computational Linguistics: EMNLP 2024}, pages 11978--11994, Miami, Florida, USA, November 2024. Association for Computational Linguistics.
\newblock \doi{10.18653/v1/2024.findings-emnlp.700}.
\newblock URL \url{https://aclanthology.org/2024.findings-emnlp.700/}.

\bibitem[Naous et~al.(2024)Naous, Ryan, Ritter, and Xu]{naous2023having}
Tarek Naous, Michael~J Ryan, Alan Ritter, and Wei Xu.
\newblock Having beer after prayer? measuring cultural bias in large language models.
\newblock In \emph{Proceedings of the 62nd Annual Meeting of the Association for Computational Linguistics (Volume 1: Long Papers)}, pages 16366--16393, 2024.

\bibitem[Ng et~al.(2024)Ng, Zhu, Song, and Xiang]{ng2024partcraft}
Kam~Woh Ng, Xiatian Zhu, Yi-Zhe Song, and Tao Xiang.
\newblock Partcraft: Crafting creative objects by parts.
\newblock In \emph{European Conference on Computer Vision}, pages 420--437. Springer, 2024.

\bibitem[OpenAI(2024)]{gpt4o}
OpenAI.
\newblock Hello gpt-4o.
\newblock News announcement by OpenAI, 2024.
\newblock URL \url{https://openai.com/index/hello-gpt-4o/}.

\bibitem[Park et~al.(2024)Park, Cho, and Ahn]{park2024mr}
Junyeong Park, Junmo Cho, and Sungjin Ahn.
\newblock Mrsteve: Instruction-following agents in minecraft with what-where-when memory.
\newblock In \emph{The Thirteenth International Conference on Learning Representations}, 2024.

\bibitem[Peng et~al.(2025)Peng, Ma, Wang, Wang, Wang, Zhang, Zhu, and Zheng]{peng2025probing}
Yongqian Peng, Yuxi Ma, Mengmeng Wang, Yuxuan Wang, Yizhou Wang, Chi Zhang, Yixin Zhu, and Zilong Zheng.
\newblock Probing and inducing combinational creativity in vision-language models.
\newblock \emph{arXiv preprint arXiv:2504.13120}, 2025.

\bibitem[Qin et~al.(2024)Qin, Zhou, Liu, Yin, Sheng, Zhang, Qiao, and Shao]{qin2024mp5}
Yiran Qin, Enshen Zhou, Qichang Liu, Zhenfei Yin, Lu~Sheng, Ruimao Zhang, Yu~Qiao, and Jing Shao.
\newblock Mp5: A multi-modal open-ended embodied system in minecraft via active perception.
\newblock In \emph{Proceedings of the IEEE/CVF Conference on Computer Vision and Pattern Recognition}, pages 16307--16316, 2024.

\bibitem[Senft-grupp et~al.(2025)Senft-grupp, Gemal, Lu, Cäsar, and Webber]{mc-bench}
Hunter Senft-grupp, Isaac Gemal, Janna Lu, Florian Cäsar, and Keith Webber.
\newblock Minecraft benchmark for large language models, 02 2025.
\newblock Manuscript in preparation.

\bibitem[Shah et~al.(2025)Shah, Kalavasis, Klivans, and Daras]{shah2025does}
Kulin Shah, Alkis Kalavasis, Adam~R Klivans, and Giannis Daras.
\newblock Does generation require memorization? creative diffusion models using ambient diffusion.
\newblock \emph{arXiv preprint arXiv:2502.21278}, 2025.

\bibitem[Shen and Guestrin()]{shen2025societal}
Judy~Hanwen Shen and Carlos Guestrin.
\newblock Societal impacts research requires benchmarks for creative composition tasks.
\newblock In \emph{ICLR 2025 Workshop on Bidirectional Human-AI Alignment}.

\bibitem[Shi et~al.(2024)Shi, Hu, Bin, Liu, Yang, Ng, Bing, and Lee]{shi2024math}
Wenhao Shi, Zhiqiang Hu, Yi~Bin, Junhua Liu, Yang Yang, See~Kiong Ng, Lidong Bing, and Roy Lee.
\newblock Math-llava: Bootstrapping mathematical reasoning for multimodal large language models.
\newblock In \emph{Findings of the Association for Computational Linguistics: EMNLP 2024}, pages 4663--4680, 2024.

\bibitem[Sorensen et~al.(2024)Sorensen, Moore, Fisher, Gordon, Mireshghallah, Rytting, Ye, Jiang, Lu, Dziri, et~al.]{sorensen2024position}
Taylor Sorensen, Jared Moore, Jillian Fisher, Mitchell Gordon, Niloofar Mireshghallah, Christopher~Michael Rytting, Andre Ye, Liwei Jiang, Ximing Lu, Nouha Dziri, et~al.
\newblock Position: a roadmap to pluralistic alignment.
\newblock In \emph{Proceedings of the 41st International Conference on Machine Learning}, pages 46280--46302, 2024.

\bibitem[Tang et~al.(2024)Tang, Song, Qin, and Yan]{tang-etal-2024-creative}
Kenan Tang, Peiyang Song, Yao Qin, and Xifeng Yan.
\newblock Creative and context-aware translation of {E}ast {A}sian idioms with {GPT}-4.
\newblock In Yaser Al-Onaizan, Mohit Bansal, and Yun-Nung Chen, editors, \emph{Findings of the Association for Computational Linguistics: EMNLP 2024}, pages 9285--9305, Miami, Florida, USA, November 2024. Association for Computational Linguistics.
\newblock \doi{10.18653/v1/2024.findings-emnlp.544}.
\newblock URL \url{https://aclanthology.org/2024.findings-emnlp.544/}.

\bibitem[Team et~al.(2023)Team, Anil, Borgeaud, Alayrac, Yu, Soricut, Schalkwyk, Dai, Hauth, Millican, et~al.]{team2023gemini}
Gemini Team, Rohan Anil, Sebastian Borgeaud, Jean-Baptiste Alayrac, Jiahui Yu, Radu Soricut, Johan Schalkwyk, Andrew~M Dai, Anja Hauth, Katie Millican, et~al.
\newblock Gemini: a family of highly capable multimodal models.
\newblock \emph{arXiv preprint arXiv:2312.11805}, 2023.

\bibitem[Tian et~al.(2024{\natexlab{a}})Tian, Huang, Liu, Jiang, Spangher, Chen, May, and Peng]{tian2024large}
Yufei Tian, Tenghao Huang, Miri Liu, Derek Jiang, Alexander Spangher, Muhao Chen, Jonathan May, and Nanyun Peng.
\newblock Are large language models capable of generating human-level narratives?
\newblock In \emph{Proceedings of the 2024 Conference on Empirical Methods in Natural Language Processing}, pages 17659--17681, 2024{\natexlab{a}}.

\bibitem[Tian et~al.(2024{\natexlab{b}})Tian, Ravichander, Qin, Le~Bras, Marjieh, Peng, Choi, Griffiths, and Brahman]{tian2023macgyver}
Yufei Tian, Abhilasha Ravichander, Lianhui Qin, Ronan Le~Bras, Raja Marjieh, Nanyun Peng, Yejin Choi, Thomas~L Griffiths, and Faeze Brahman.
\newblock Macgyver: Are large language models creative problem solvers?
\newblock In \emph{Proceedings of the 2024 Conference of the North American Chapter of the Association for Computational Linguistics: Human Language Technologies (Volume 1: Long Papers)}, pages 5303--5324, 2024{\natexlab{b}}.

\bibitem[Wang et~al.(2024{\natexlab{a}})Wang, Huang, Ma, Song, Lu, Bian, Li, Liu, and Li]{wang2024zola}
Fu-Yun Wang, Zhaoyang Huang, Qiang Ma, Guanglu Song, Xudong Lu, Weikang Bian, Yijin Li, Yu~Liu, and Hongsheng Li.
\newblock Zola: Zero-shot creative long animation generation with short video model.
\newblock In \emph{European Conference on Computer Vision}, pages 329--345. Springer, 2024{\natexlab{a}}.

\bibitem[Wang et~al.(2024{\natexlab{b}})Wang, Pan, Shi, Lu, Ren, Zhou, Zhan, and Li]{wang2024measuring}
Ke~Wang, Junting Pan, Weikang Shi, Zimu Lu, Houxing Ren, Aojun Zhou, Mingjie Zhan, and Hongsheng Li.
\newblock Measuring multimodal mathematical reasoning with math-vision dataset.
\newblock \emph{Advances in Neural Information Processing Systems}, 37:\penalty0 95095--95169, 2024{\natexlab{b}}.

\bibitem[Wang et~al.(2024{\natexlab{c}})Wang, Liu, and Lv]{wang2024create}
Letian Wang, Xianggen Liu, and Jiancheng Lv.
\newblock Create! don’t repeat: A paradigm shift in multi-label augmentation through label creative generation.
\newblock In \emph{Proceedings of the 2024 Conference of the North American Chapter of the Association for Computational Linguistics: Human Language Technologies (Volume 1: Long Papers)}, pages 855--869, 2024{\natexlab{c}}.

\bibitem[Wang et~al.(2023{\natexlab{a}})Wang, Que, Chen, Li, Li, and Yang]{wang2023creative}
Renke Wang, Guimin Que, Shuo Chen, Xiang Li, Jun Li, and Jian Yang.
\newblock Creative birds: self-supervised single-view 3d style transfer.
\newblock In \emph{Proceedings of the IEEE/CVF international conference on computer vision}, pages 8775--8784, 2023{\natexlab{a}}.

\bibitem[Wang et~al.(2024{\natexlab{d}})Wang, Chen, Wang, Cao, Liu, Gao, Zhu, Zhu, Lu, Qiao, et~al.]{wang2024enhancing}
Weiyun Wang, Zhe Chen, Wenhai Wang, Yue Cao, Yangzhou Liu, Zhangwei Gao, Jinguo Zhu, Xizhou Zhu, Lewei Lu, Yu~Qiao, et~al.
\newblock Enhancing the reasoning ability of multimodal large language models via mixed preference optimization.
\newblock \emph{arXiv preprint arXiv:2411.10442}, 2024{\natexlab{d}}.

\bibitem[Wang et~al.(2023{\natexlab{b}})Wang, Wei, Schuurmans, Le, Chi, Narang, Chowdhery, and Zhou]{wang2022self}
Xuezhi Wang, Jason Wei, Dale Schuurmans, Quoc~V. Le, Ed~H. Chi, Sharan Narang, Aakanksha Chowdhery, and Denny Zhou.
\newblock Self-consistency improves chain of thought reasoning in language models.
\newblock In \emph{The Eleventh International Conference on Learning Representations, {ICLR} 2023, Kigali, Rwanda, May 1-5, 2023}, 2023{\natexlab{b}}.

\bibitem[Wang et~al.(2024{\natexlab{e}})Wang, Xia, He, Chen, Liu, Zhu, Liang, Wu, Liu, Malladi, et~al.]{wang2024charxiv}
Zirui Wang, Mengzhou Xia, Luxi He, Howard Chen, Yitao Liu, Richard Zhu, Kaiqu Liang, Xindi Wu, Haotian Liu, Sadhika Malladi, et~al.
\newblock Charxiv: Charting gaps in realistic chart understanding in multimodal llms.
\newblock \emph{Advances in Neural Information Processing Systems}, 37:\penalty0 113569--113697, 2024{\natexlab{e}}.

\bibitem[White et~al.(2025)White, Nottingham, Maniar, Robinson, Lillemark, Maheshwari, Qin, and Ammanabrolu]{white2025collaborating}
Isadora White, Kolby Nottingham, Ayush Maniar, Max Robinson, Hansen Lillemark, Mehul Maheshwari, Lianhui Qin, and Prithviraj Ammanabrolu.
\newblock Collaborating action by action: A multi-agent llm framework for embodied reasoning.
\newblock \emph{arXiv preprint arXiv:2504.17950}, 2025.

\bibitem[Yao et~al.(2024)Yao, Yu, Zhang, Wang, Cui, Zhu, Cai, Li, Zhao, He, et~al.]{yao2024minicpm}
Yuan Yao, Tianyu Yu, Ao~Zhang, Chongyi Wang, Junbo Cui, Hongji Zhu, Tianchi Cai, Haoyu Li, Weilin Zhao, Zhihui He, et~al.
\newblock Minicpm-v: A gpt-4v level mllm on your phone.
\newblock \emph{arXiv preprint arXiv:2408.01800}, 2024.

\bibitem[Yue et~al.(2024)Yue, Ni, Zhang, Zheng, Liu, Zhang, Stevens, Jiang, Ren, Sun, et~al.]{yue2024mmmu}
Xiang Yue, Yuansheng Ni, Kai Zhang, Tianyu Zheng, Ruoqi Liu, Ge~Zhang, Samuel Stevens, Dongfu Jiang, Weiming Ren, Yuxuan Sun, et~al.
\newblock Mmmu: A massive multi-discipline multimodal understanding and reasoning benchmark for expert agi.
\newblock In \emph{Proceedings of the IEEE/CVF Conference on Computer Vision and Pattern Recognition}, pages 9556--9567, 2024.

\bibitem[Zhang et~al.(2024)Zhang, Chen, Zhao, Chen, Tang, and Liang]{zhang2024hidiffusion}
Shen Zhang, Zhaowei Chen, Zhenyu Zhao, Yuhao Chen, Yao Tang, and Jiajun Liang.
\newblock Hidiffusion: Unlocking higher-resolution creativity and efficiency in pretrained diffusion models.
\newblock In \emph{European Conference on Computer Vision}, pages 145--161. Springer, 2024.

\bibitem[Zhang et~al.(2025)Zhang, Hu, Lee, Shi, Kordjamshidi, Chai, and Ma]{zhang2025do}
Zheyuan Zhang, Fengyuan Hu, Jayjun Lee, Freda Shi, Parisa Kordjamshidi, Joyce Chai, and Ziqiao Ma.
\newblock Do vision-language models represent space and how? evaluating spatial frame of reference under ambiguities.
\newblock In \emph{The Thirteenth International Conference on Learning Representations}, 2025.
\newblock URL \url{https://openreview.net/forum?id=84pDoCD4lH}.

\bibitem[Zhong et~al.(2024)Zhong, Huang, Gao, Wen, Lin, Zitnik, and Zhou]{zhong2024let}
Shanshan Zhong, Zhongzhan Huang, Shanghua Gao, Wushao Wen, Liang Lin, Marinka Zitnik, and Pan Zhou.
\newblock Let's think outside the box: Exploring leap-of-thought in large language models with creative humor generation.
\newblock In \emph{Proceedings of the IEEE/CVF Conference on Computer Vision and Pattern Recognition}, pages 13246--13257, 2024.

\bibitem[Zhu et~al.(2024)Zhu, Huang, Rudinac, and Kanoulas]{zhu2024enhancing}
Hongyi Zhu, Jia-Hong Huang, Stevan Rudinac, and Evangelos Kanoulas.
\newblock Enhancing interactive image retrieval with query rewriting using large language models and vision language models.
\newblock In \emph{Proceedings of the 2024 International Conference on Multimedia Retrieval}, pages 978--987, 2024.

\bibitem[Zhu et~al.(2025)Zhu, Jia, Zhang, Li, and Jiang]{zhu2024multichartqa}
Zifeng Zhu, Mengzhao Jia, Zhihan Zhang, Lang Li, and Meng Jiang.
\newblock {M}ulti{C}hart{QA}: Benchmarking vision-language models on multi-chart problems.
\newblock In \emph{Proceedings of the 2025 Conference of the Nations of the Americas Chapter of the Association for Computational Linguistics: Human Language Technologies (Volume 1: Long Papers)}, pages 11341--11359, Albuquerque, New Mexico, April 2025.
\newblock ISBN 979-8-89176-189-6.

\end{thebibliography}
\bibliographystyle{plainnat}


\appendix



\newpage

\section{Limitations}
\label{Appendix:limitatitons}

There are two primary limitations. First, due to disparities in technological development across countries and regions, the majority of players in the ``Guess the Build'' game on the Hypixel server are from more developed areas. As a result, the data we collect lacks balanced geographical representation at the global level. Second, in our dynamic task setting, current VLMs do not yet support directly answering questions based on video input. Therefore, we simulate the dynamic process using three static images accompanied by corresponding hints. We anticipate that both limitations can be addressed in the future as global technological infrastructure improves and any-to-any models become available.

\section{Ethics Statement}
\label{Appendix:ethics_statement}

\begin{figure}[t]
    \centering
    \includegraphics[width=\linewidth]{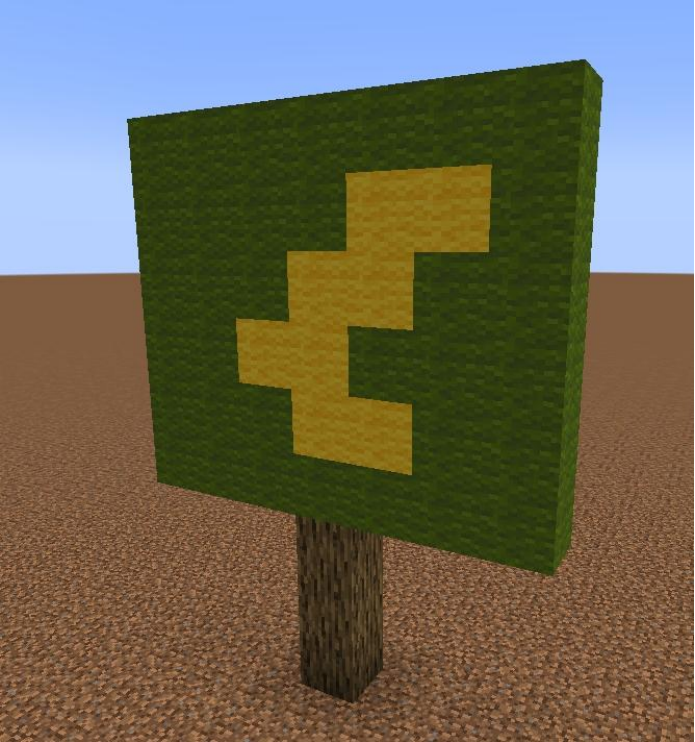}
    \vspace{-15pt}
    \caption{A Minecraft build depicting the China Post logo.}
    \label{fig:chinese_post}
\end{figure}

Due to the presence of social and cultural biases~\cite{naous2023having}, VLMs tend to exhibit higher accuracy when predicting certain concepts over others, and their interpretations often align more closely with Western cultural contexts. For example, as shown in Figure~\ref{fig:chinese_post}, the Minecraft build represents the logo of China Post. While players familiar with this symbol can easily identify it, GPT-4o describes it as ``fast food.'' This example illustrates that VLMs exhibit a clear cultural bias in their understanding of creative content.

\section{Experiment Details}
\label{appendix:exp_details}

\subsection{Instructions}

Table~\ref{tab:appendix_instructions} presents the default instructions used in \ourdataset{}. For the static task, where only a single guess is allowed, we use the \textit{Initial Guess Instruction} exclusively. For the dynamic task, the \textit{Initial Guess Instruction} is used for the first attempt. If the response is incorrect, subsequent attempts incorporate the model’s previous guess to construct the \textit{Guess Refinement Instruction}.

\begin{table}[th!]
  \centering
    \scalebox{0.92}{
    \begin{tabular}{p{3cm}p{11cm}}
    
    \hline
    \toprule
    \textbf{Category} & \textbf{Detail} \\
    \midrule
    
    System & You are an AI designed to play the ``Guess the Build'' game from the Hypixel server in Minecraft. Your task is to analyze a player's build and accurately guess what it represents. The build is constructed using Minecraft blocks, and it can depict objects, animals, structures, abstract concepts, or any other recognizable entity.

    Consider common themes, contextual clues, and typical constructions in Minecraft. If a build is unclear, make educated guesses based on possible interpretations. Your goal is to guess as accurately and efficiently as possible.
    
    Avoid overly broad answers—be specific when possible. If multiple plausible answers exist, prioritize the most likely one based on common builds in the game.
    \\
    
    \midrule
        
    Initial Guess  & Look at the provided image of a Minecraft build and the corresponding hint. Based on its structure, shape, materials, and the given hint, determine what it represents. Consider common objects, animals, buildings, abstract concepts, or any other recognizable entities typically built in Minecraft.

    You can use chain of thought reasoning to arrive at the best possible guess. \\
    
    \midrule
        
    Guess Refinement & Your previous guess is: \textit{previous\_guess}. This guess is incorrect. Now, you will be provided with a more complete version of the build along with additional hints. Analyze the new information carefully and make a revised guess based on the updated build and hints.

    You can use chain of thought reasoning to improve your guess. \\
    
    \bottomrule
    \hline
    \end{tabular}
    }
    \vspace{1ex}
    \caption{Illustrations of Instructions. The table presents the default instructions used for evaluating \ourdataset{}. During evaluation, either the \textit{Initial Guess Instruction} or the \textit{Guess Refinement Instruction} is concatenated with the corresponding hint and then provided as input to the VLM.}
    \label{tab:appendix_instructions}
\end{table}

\subsection{Answer Extraction Prompt}
\label{Appendix:ans_extractor}

Table~\ref{tab:appendix_prompt_for_answer_extraction} presents the instruction provided to GPT-4o for extracting the predicted guess generated by the VLMs.

\begin{table}[th!]
  \centering
    \scalebox{1.0}{
    \begin{tabular}{p{12cm}}
    
    \hline
    \toprule
    \textbf{Instruction}  \\
    \midrule

    I will give you a language model's response in a `Guess the Build' game. Identify and extract the model's guessed answer. Output it in the format: `Answer: [extracted answer]'. If no answer is given in the response, output `No Answering'. \\

    Language model's response:  \\

    \bottomrule
    \hline
    \end{tabular}
    }
    \vspace{1ex}
    \caption{Illustrations of the instructions used for guess extraction. The response from the VLM is appended directly after these instructions, and the combined input is then fed into GPT-4o to extract the predicted guess.}
    \label{tab:appendix_prompt_for_answer_extraction}
\end{table}

\subsection{Prompt for Different Reasoning Approaches}
\label{Appendix:reasoning_approaches}

Table~\ref{tab:appendix_reasoning_approach_instruction} presents the core instructions for the different reasoning approaches. For the Self-Consistency method, no specific instruction is provided; instead, the VLM is prompted three times, and the final answer is determined via majority voting. 

\begin{table}[th!]
  \centering
    \scalebox{0.92}{
    \begin{tabular}{p{3cm}p{11cm}}
    
    \hline
    \toprule
    \textbf{Approach} & \textbf{Detail} \\
    \midrule
    
    Without CoT & You are not allowed to use chain-of-thought reasoning. You must output the final answer directly.
    \\
    
    \midrule
        
    One-shot  & I will give you an example to help you guess the build. Here is an example:\newline\newline
    <Image>\newline
    Hint\_3: The answer format is as follows: \newline
    \phantom{aa}l\phantom{aaaa}s\phantom{aaaaaaaaaaaa}k\newline
    \_ \_ \_ \_ \_\phantom{aa}\_ \_\phantom{aa}\_ \_ \_ \_\newline
    This means the answer consists of 3 words. The 1st word has 5 letters. The 2nd word has 2 letters. The 3rd word has 4 letters. The 1st word's 2nd letter is l. The 1st word's 5th letter is s. The 3rd word's 4th letter is k.\newline
    Answer: The build represents Glass of Milk.\\
    
    \midrule
        
    Self-Refine & Here is your response: \textit{previous\-response}
    
    Please review your answer to check if it aligns with the given hint. If it does, you MUST ONLY output ``\$Well done!\$" Otherwise, provide your improved guess.\\

    \midrule

    Image Retrieval & Previously, your guess was: \textit{previous\_guess}

    Now, I have searched for a real image of \textit{previous\_guess} from Google Images and attached it above. Please compare the provided Minecraft build with the real image of your guess.

    If they match, you MUST ONLY output: ``\$Well done!\$". Otherwise, provide your improved guess.\\
    
    \bottomrule
    \hline
    \end{tabular}
    }
    \vspace{1ex}
    \caption{Illustrations of the instructions used for different reasoning approaches. For each reasoning approach, we present the core instruction to provide a more intuitive and accessible understanding.}
    \label{tab:appendix_reasoning_approach_instruction}
\end{table}

\subsection{Model Hyperparameters}

Table~\ref{tab:appendix_VLM_generating_para} presents the detailed parameter configurations for the models discussed in (\S\ref{3.1:models_and_implementation}). Parameters not explicitly specified follow their default settings. Notably, for the Self-Consistency experiments described in (\S\ref{3.2:reasoning_approach}), we allow sampling and do not manually set the temperature or top\_p parameters, instead relying on their default values. This design choice is intended to promote response diversity during generation, thereby ensuring that the results obtained through majority voting are meaningful.

\begin{table}[th!]
\centering
\begin{tabular}{p{0.2\linewidth} | p{0.7\linewidth}}
\toprule
\textbf{Model} & \textbf{Generation Setup} \\
\midrule
 GPT-4o & model = \texttt{gpt-4o-2024-11-20}, temperature = 0.0, top\_p=1.0, max\_tokens = 1000 \\
\midrule 
 Gemini-2.0-Flash & model = \texttt{gemini-2.0-flash-001}, temperature = 0.0, top\_p=1.0, max\_output\_tokens=1000 \\
\midrule
GPT-4o-mini &  model = \texttt{gpt-4o-mini-2024-07-18}, temperature = 0.0, top\_p=1.0, max\_tokens = 1000 \\
\midrule
 InternVL2.5-MPO & max\_new\_tokens = 1000, do\_sample = False \\
\midrule
Qwen2.5VL-72B & max\_new\_tokens = 1000, do\_sample = False \\
\midrule
 MiniCPM-V2.6 &  max\_new\_tokens = 1000, sampling = False \\

\bottomrule
\end{tabular}
\vspace{1ex}
\caption{Generating parameters for VLMs.}
\label{tab:appendix_VLM_generating_para}
\end{table}

\subsection{Implementation Details for Fine-tuning}
\label{appendix:fine-tune_details}

We use the Unsloth framework~\cite{unsloth} and apply a consistent set of hyperparameters for the three fine-tuning settings: without tuning, synthetic tuning, and mixed tuning, as shown in Table~\ref{tab:appendix_fine-tuning_para}. The fine-tuning process is performed using a single NVIDIA A100-SXM4-40GB GPU.

\begin{table}[th!]
\centering
\begin{tabular}{p{0.13\linewidth} | p{0.77\linewidth}}
\toprule

\textbf{Category} & \textbf{Detail} \\
\midrule

Model & model\_name = ``unsloth/Qwen2-VL-7B-Instruct'', load\_in\_4bit = True \\

\midrule 

LoRA & r = 16, lora\_alpha = 16, lora\_dropout = 0, random\_state = 3407,\\

\midrule

Fine-tuning  & per\_device\_train\_batch\_size = 2, gradient\_accumulation\_steps = 4, warmup\_steps = 5, max\_steps = 30, learning\_rate = 2e-4, \newline 
optim = ``adamw\_8bit'', weight\_decay = 0.01, lr\_scheduler\_type = ``linear'',\newline 
seed = 3407, max\_seq\_length = 2048,\\

\bottomrule
\end{tabular}
\vspace{1ex}
\caption{Fine-tuning Configuration and Hyperparameters.}
\label{tab:appendix_fine-tuning_para}
\end{table}

\subsection{Performance of Selected Models on Existing Benchmarks}
\label{appendix:detailed_accuracy}

We obtained the accuracy of all selected models on the four existing datasets from the OpenVLM Leaderboard\footnote{\url{https://huggingface.co/spaces/opencompass/open_vlm_leaderboard}}. To ensure consistency with the evaluation metric used in \ourdataset{}, which is the average accuracy per question, we selected the \textit{Overall Accuracy} metric for the AI2D, MMBench V1.1 Test (EN), and MathVista benchmarks. For HallusionBench, we instead used \textit{aAcc}, which represents the overall accuracy across all atomic questions.

\begin{table*}[th!]
\resizebox{1\linewidth}{!}{
      \begin{tabular}{lcccccc@{}}
        \hline
        \toprule
        \textbf{Dataset} & \phantom{} &
        \textbf{GPT-4o} & 
        \textbf{Gemini-2.0-Flash} & 
        \textbf{InternVL2.5-78B-MPO} &
        \textbf{Qwen2.5VL-72B} &
        \textbf{MiniCPM-V2.6} \\

        \midrule

        AI2D &  & 84.9 & 83.1 & 89.2 & 88.5 & 82.1 \\
        
        MMBench V1.1 Test(EN) &  & 84.8 & 70.4 & 87.9 & 88.3 & 79.0 \\
        
        MathVista &  & 60.0 & 70.4 & 76.6 & 74.2 & 60.8 \\

        HallusionBench &  & 71.4 & 72.0 & 73.3 & 71.9 & 65.0 \\

        \bottomrule
        \hline
      \end{tabular}
    }
    \caption{The detailed accuracy of all models on the four existing datasets.}
  \vspace{-2ex}
  \label{appendix:tab_detailed_acc}
\end{table*}

\section{Sycophantic Behavior in VLMs under Contradictory User Feedback}

To investigate whether VLMs exhibit sycophantic behavior when confronted with contradictory user requests, we conduct a study focusing on two state-of-the-art models: GPT-4o and Gemini-2.0-Flash. Within the static evaluation setting, we select 100 instances from the Minecraft build sets that both models originally answered correctly. For each instance, after the VLM provides a correct response, we intentionally assert that its answer is incorrect, allowing us to observe whether the model revises its correct prediction and instead conforms to the user's false feedback.

We design three conditions to induce such behavior in VLMs. In the \textbf{Base} setting, we merely inform the VLM that its answer is wrong without providing any additional information. The \textbf{Random} setting builds upon the Base condition by additionally supplying a randomly chosen incorrect answer, sampled from the set of incorrect options in \ourdataset{}, and asserting it as correct. The \textbf{Similar} setting also builds upon the Base condition, but the misleading answer is semantically similar to the correct one, generated using GPT-4o, and presented as the correct answer.

The results reveal that GPT-4o maintains an accuracy of 6\%, 15\%, and 26\% under the Base, Random, and Similar settings respectively. In contrast, Gemini-2.0-Flash achieves only 1\%, 2\%, and 4\% accuracy across the same conditions. These findings suggest that Gemini-2.0-Flash is more susceptible to sycophantic behavior than GPT-4o.

Interestingly, VLMs are most likely to revise their initially correct answers in the Base setting, where no alternative answer is proposed, and are least likely to do so in the Similar setting, where a plausible but incorrect alternative is given. Further analysis of the model responses under contradictory user feedback reveals that in both the Random and Similar settings, the VLMs do not engage in any substantive evaluation of the user-supplied answer’s plausibility. Instead, they often respond with generic phrases such as “reconsidering the build and the hint,” immediately followed by a new, often incorrect, guess.

Paradoxically, as the misleading input becomes more semantically aligned with the correct answer from Base to Random to Similar, the models’ accuracy increases. This trend runs counter to our hypothesis. If the models were capable of independent reasoning, misleading cues that closely resemble the correct answer would be more deceptive and thus reduce accuracy. Instead, our results suggest that VLMs fundamentally lack the ability to assess correctness or engage in independent thought. Rather, they merely generate text that appears syntactically plausible in response to user prompts.

\section{Flipping the Paradigm: Easier to Generate than Discriminate}

Existing experiments show that generative tasks are generally more challenging for LMs than discriminative ones \citep{jiang2025self}. However, our findings reveal a surprising result: creatively generating a Minecraft build is significantly easier than identifying one, further underscoring the novelty and distinctiveness of \ourdataset{}.

In our generative experiment setup, we select the Gemini-2.0-Flash model and randomly sample 100 build sets from \ourdataset{}. For each sampled item, we manually input prompts into the Gemini playground\footnote{\url{https://aistudio.google.com/prompts/new_chat}}, asking the model to generate an image of a Minecraft build based on a given answer, thereby reversing the question-answering direction in \ourdataset{}. For the discriminative counterpart, we adopt the \textit{static task} from \ourdataset{} as our evaluation setting.

We evaluate performance using accuracy. A generation is considered correct only if it satisfies two criteria: it contains the specified entities, and it appears plausibly constructed within the Minecraft environment. Under this rigorous standard, Gemini-2.0-Flash achieves an accuracy of 80\% on the generative task, which is substantially higher than its 40\% accuracy on the corresponding discriminative task within \ourdataset{}.

This notable performance gap highlights a critical limitation in VLMs' ability to perform creative understanding in open-ended scenarios, a challenge that remains largely overlooked in prior work. \ourdataset{} not only exposes this limitation but also points to promising directions for future improvements in VLM capabilities.

\section{From Feedback to Foresight: GPT-4o Learns in Context Like a Player}

To more accurately simulate the gameplay of ``Guess the Build'' and evaluate the performance of VLMs in a realistic gaming environment, we divide a total of 500 questions into 100 sets, each containing 5 questions. This setup mirrors the actual game mechanics, where five questions are asked consecutively, and after each question, the VLM receives feedback indicating whether its guess is correct or incorrect. In our experiments, we adopt GPT-4o as the VLM under investigation.

\begin{figure}[t]
    \centering
    \includegraphics[width=\linewidth]{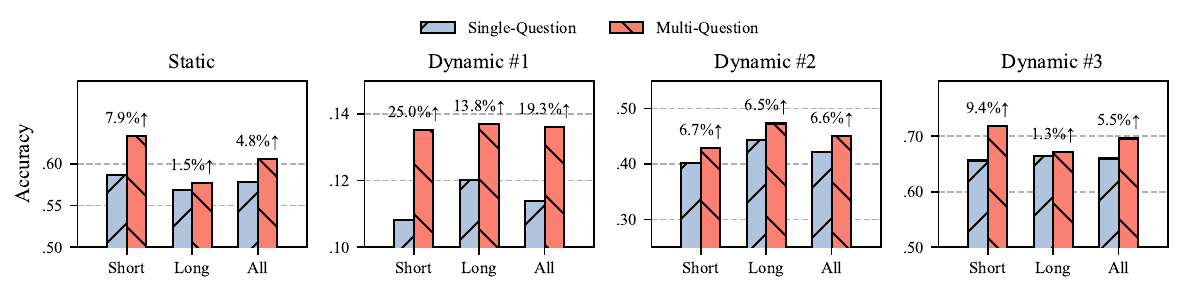}
    \vspace{-15pt}
    \caption{Evaluation results of Single-Question and Multi-Question settings. All denotes the overall accuracy for each task. Answering multiple questions sequentially within the same context improves response accuracy.}
    \label{fig: multi-questions}
\end{figure}

As shown in Figure~\ref{fig: multi-questions}, answering five questions in succession leads to improved accuracy, with a more substantial performance gain observed in dynamic tasks. Specifically, accuracy increases by 4.8\% in static tasks and by an average of 10.47\% across three attempts in dynamic tasks. These results indicate that GPT-4o is able to extract more information from dynamic tasks and utilize prior interactions through in-context learning to enhance its responses to subsequent questions. Moreover, we observe that in dynamic tasks, the improvement in accuracy diminishes with each additional attempt. This suggests that the benefits of GPT-4o's in-context learning are more pronounced when dealing with more incomplete Minecraft builds and less informative hints.

This experiment demonstrates that GPT-4o possesses strong in-context learning capabilities, enabling it to answer new questions more effectively by drawing on previous ones, similar to human players. It also highlights a promising direction for enhancing VLMs' ability to understand creativity in real-world contexts.

\section{Reasoning Reliability Analysis}

To evaluate whether VLMs can accurately explain their reasoning path when correctly identifying a Minecraft build, we conduct a manual analysis under the static task setting. Specifically, we randomly sample 50 responses from all correctly answered questions by GPT-4o and examine the reasoning path provided in each response. We find that 14\% of the responses contain issues in their reasoning paths. Among these, 42.8\% contain flawed reasoning (for example, inferring that a build resembles ``a pit or a mine with ores inside'' based on appearance, but ultimately identifying it as a ``crate''), 28.6\% involve errors in visual perception (such as stating that a table has four legs when the image clearly shows only three), and the remaining 28.6\% result from inaccurate textual perception (for instance, misinterpreting the hint ``\_ l a \_ \_ \_ \_'' as ``l a''). These findings suggest that GPT-4o occasionally generates incorrect reasoning even when the final answer is accurate, underscoring its limitations in creativity understanding.

\section{Examples}

\subsection{Illustrative Examples for Static and Dynamic Tasks}
\label{appendix:static_dynamic_eg}

For both the \emph{static} and \emph{dynamic} task settings, we selected the example ``Graveyard'' to illustrate the entire process within each task, including how GPT-4o receives the question, produces a response, and is subsequently evaluated for correctness. This example is intended to help readers gain a clearer understanding of the specific nature and structure of the two tasks.

\begin{figure}[t]
    \centering
    \includegraphics[width=1\linewidth]{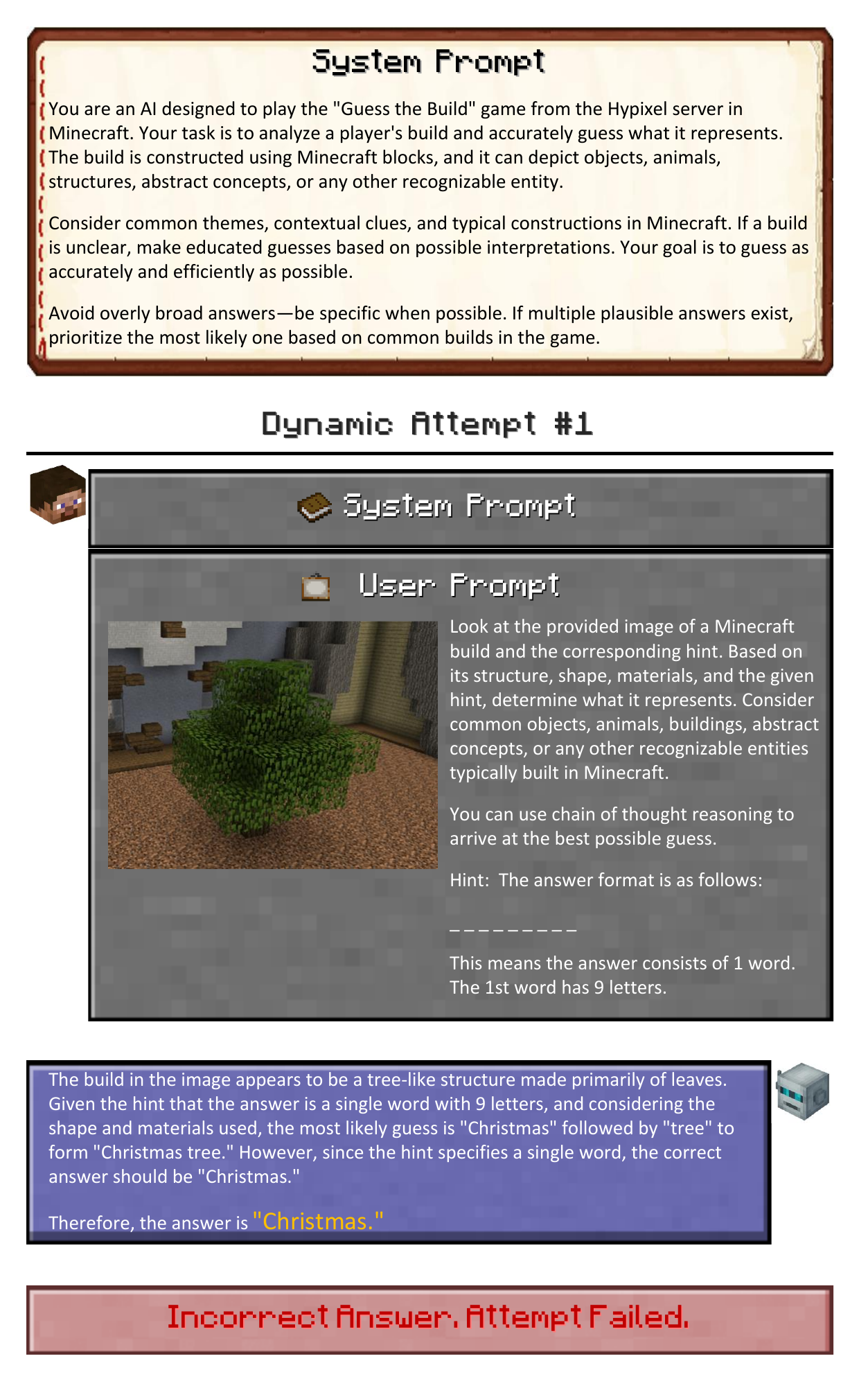}
    \caption{An example illustrates the first attempt in the dynamic task setting: GPT-4o responds based on an incomplete Minecraft build and a partial hint, but produces an incorrect answer.}
    \label{fig: appendix_dynamic1}
\end{figure}

\begin{figure}[t]
    \centering
    \includegraphics[width=1\linewidth]{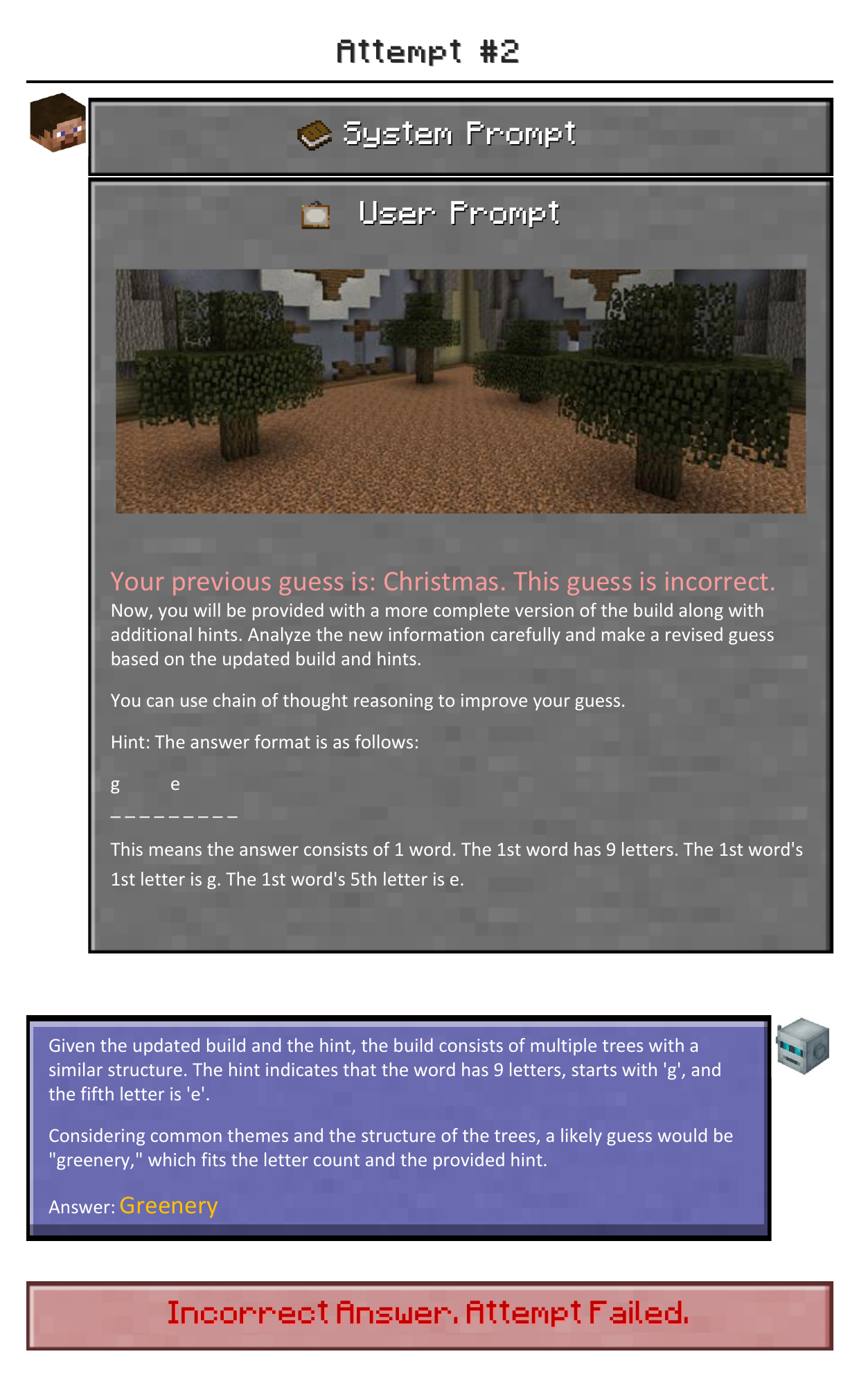}
    \vspace{-25pt}
    \caption{An example illustrates the second attempt in the dynamic task setting: GPT-4o, with access to the details of the first attempt, responds again based on a more complete Minecraft build and a more informative hint, but still produces an incorrect answer.}
    \label{fig: appendix_dynamic2}
\end{figure}

\begin{figure}[t]
    \centering
    \includegraphics[width=1\linewidth]{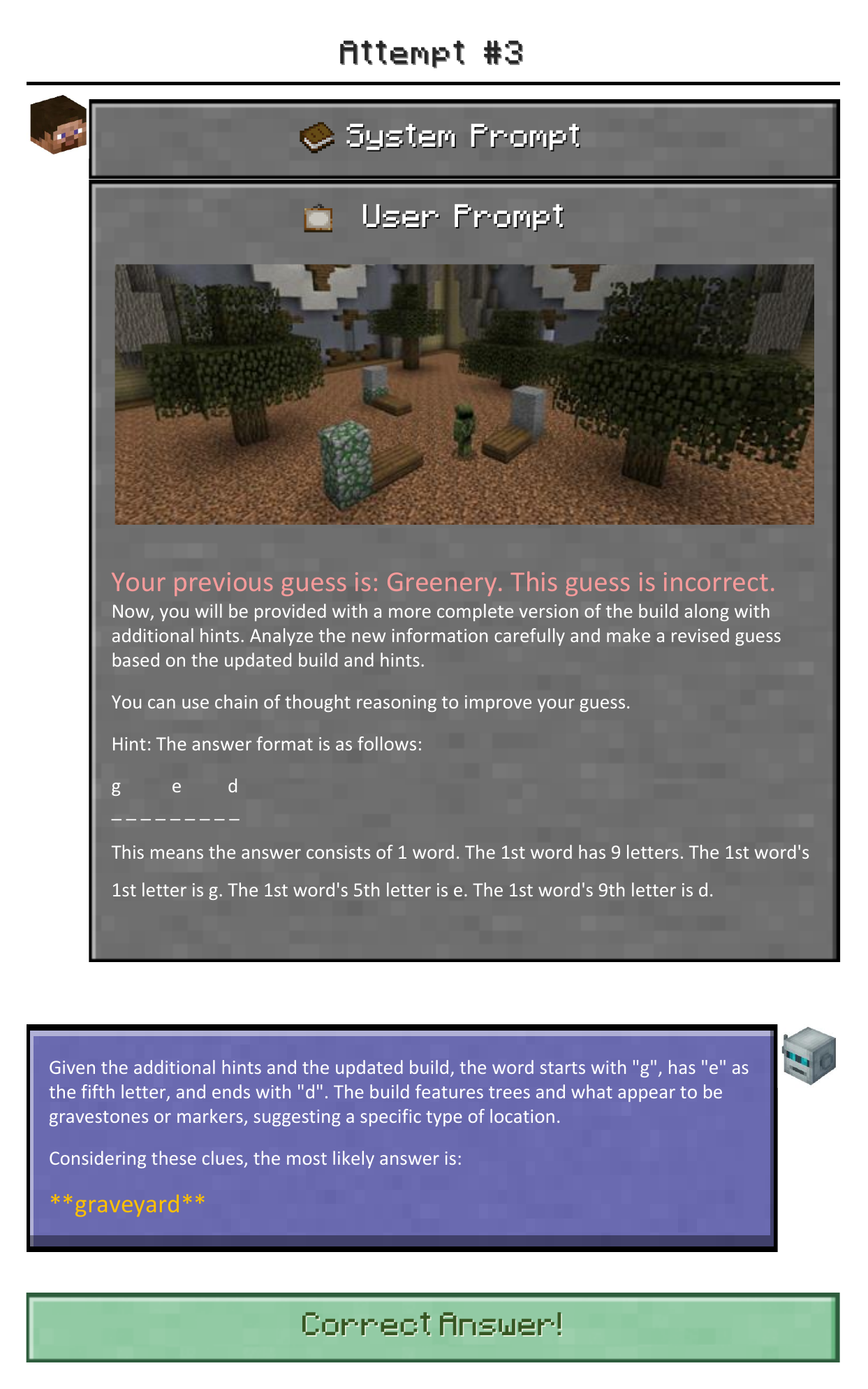}
    \vspace{-23pt}
    \caption{An example illustrates the third attempt in the dynamic task setting: GPT-4o, with access to the details of the previous two attempts, responds again based on the most complete Minecraft build and the most comprehensive hint, and produces a correct answer.}
    \label{fig: appendix_dynamic3}
\end{figure}

\begin{figure}[t]
    \centering
    \includegraphics[width=1\linewidth]{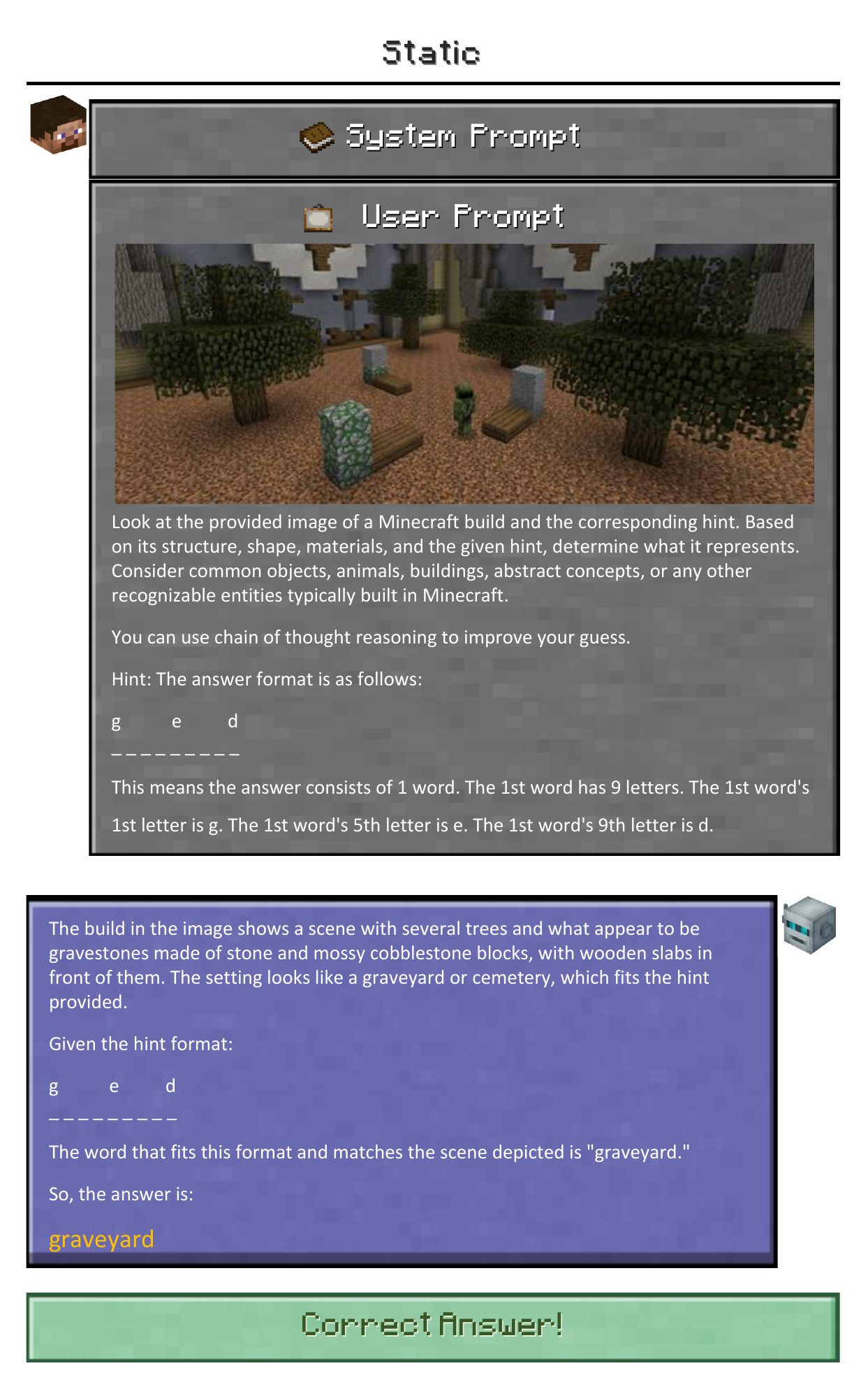}
    \vspace{-20pt}
    \caption{An example of the static task setting: GPT-4o, based on a complete Minecraft build and a full hint, makes a prediction and provides the correct answer.}
    \label{fig: appendix_static}
\end{figure}

\subsection{Case Study: Image Retrieval Reduces Answer Accuracy}
\label{appendix:image_retrieval_fail}

We present a case where the use of an \textbf{Image Retrieval} method unexpectedly reduces the accuracy of the model’s prediction. Initially, GPT-4o correctly predicts the answer as \emph{Modern Art}. However, since \emph{Modern Art} encompasses a wide range of architectural styles and painting techniques, the image retrieval method returns an image of \emph{Starry Night} by Vincent van Gogh. Although this painting falls under the category of \emph{Modern Art}, it differs significantly from the original Minecraft build in visual appearance. As a result, GPT-4o alters its originally correct prediction. Subsequent image retrievals and comparisons fail to guide the model back to the correct answer.

\begin{figure}[t]
    \centering
    \includegraphics[width=1\linewidth]{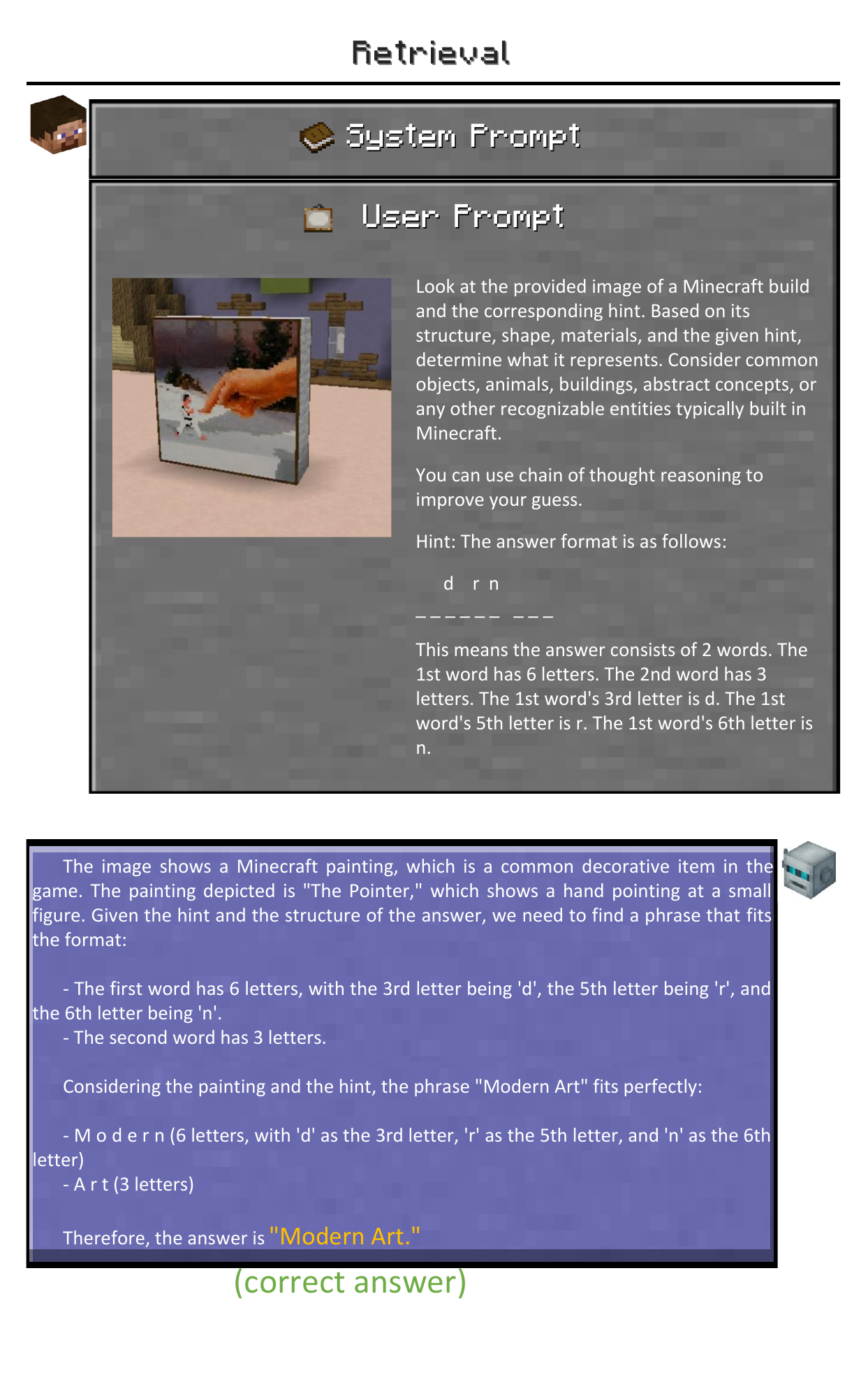}
    \vspace{-35pt}
    \caption{One example from the static task. Initially, GPT-4o provides the correct answer based on the Minecraft build and the accompanying hint.}
    \label{fig: appendix_retrieval1}
\end{figure}

\begin{figure}[t]
    \centering
    \includegraphics[width=1\linewidth]{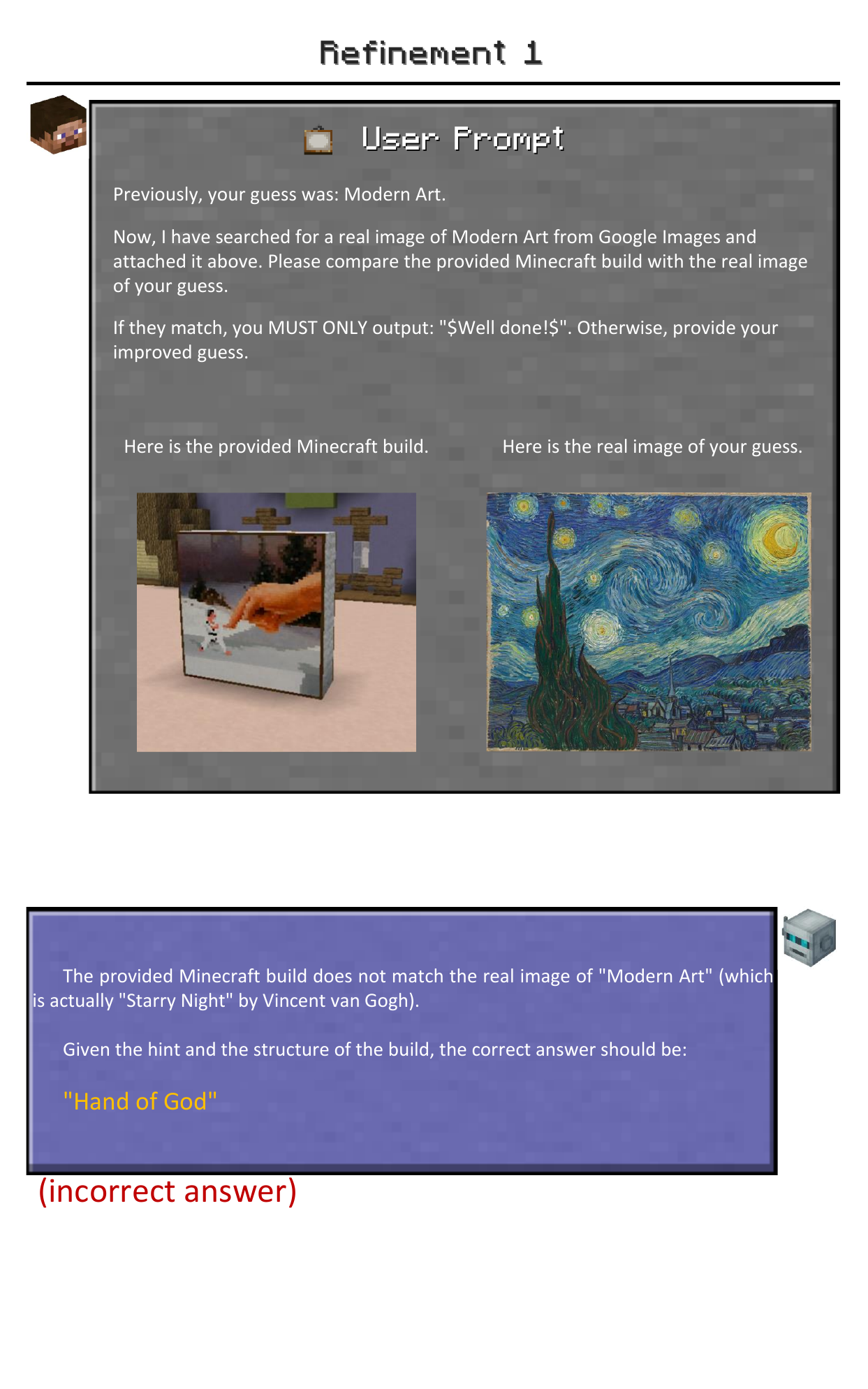}
    \vspace{-60pt}
    \caption{Following the results of the first image retrieval, GPT-4o revises its initially accurate prediction after retrieving and comparing the images, resulting in an incorrect response. At this stage, GPT-4o remains in the refinement process and has not yet concluded, as it has not undergone evaluation.}
    \label{fig: appendix_retrieval2}
\end{figure}

\begin{figure}[t]
    \centering
    \includegraphics[width=1\linewidth]{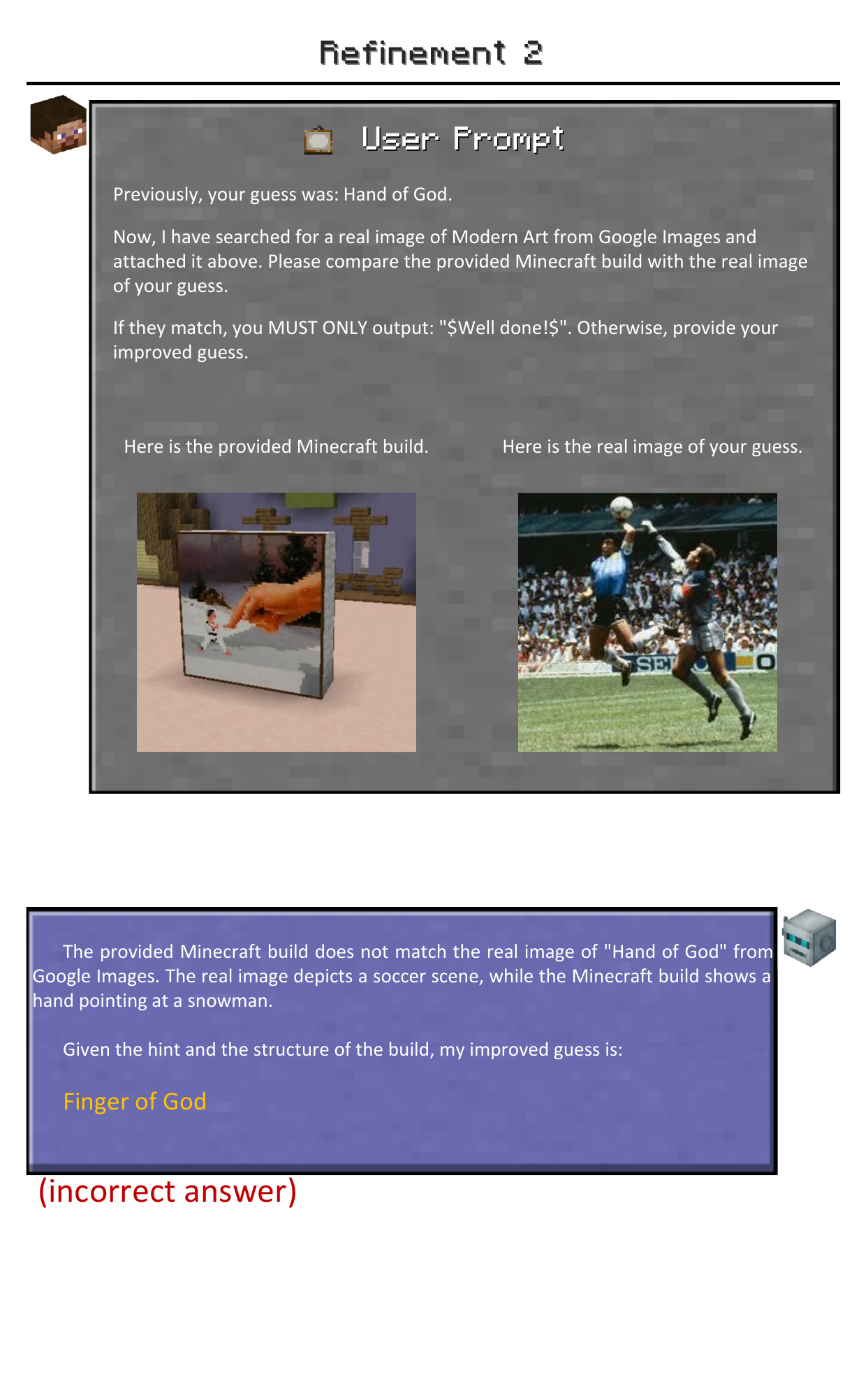}
    \vspace{-60pt}
    \caption{Subsequently, GPT-4o performs a second round of image retrieval. After retrieving and comparing the images, it again modifies its original prediction; however, the response remains incorrect. At this point, GPT-4o is still in the refinement process and has not yet reached completion, as it has not undergone final evaluation.}
    \label{fig: appendix_retrieval3}
\end{figure}

\begin{figure}[t]
    \centering
    \includegraphics[width=1\linewidth]{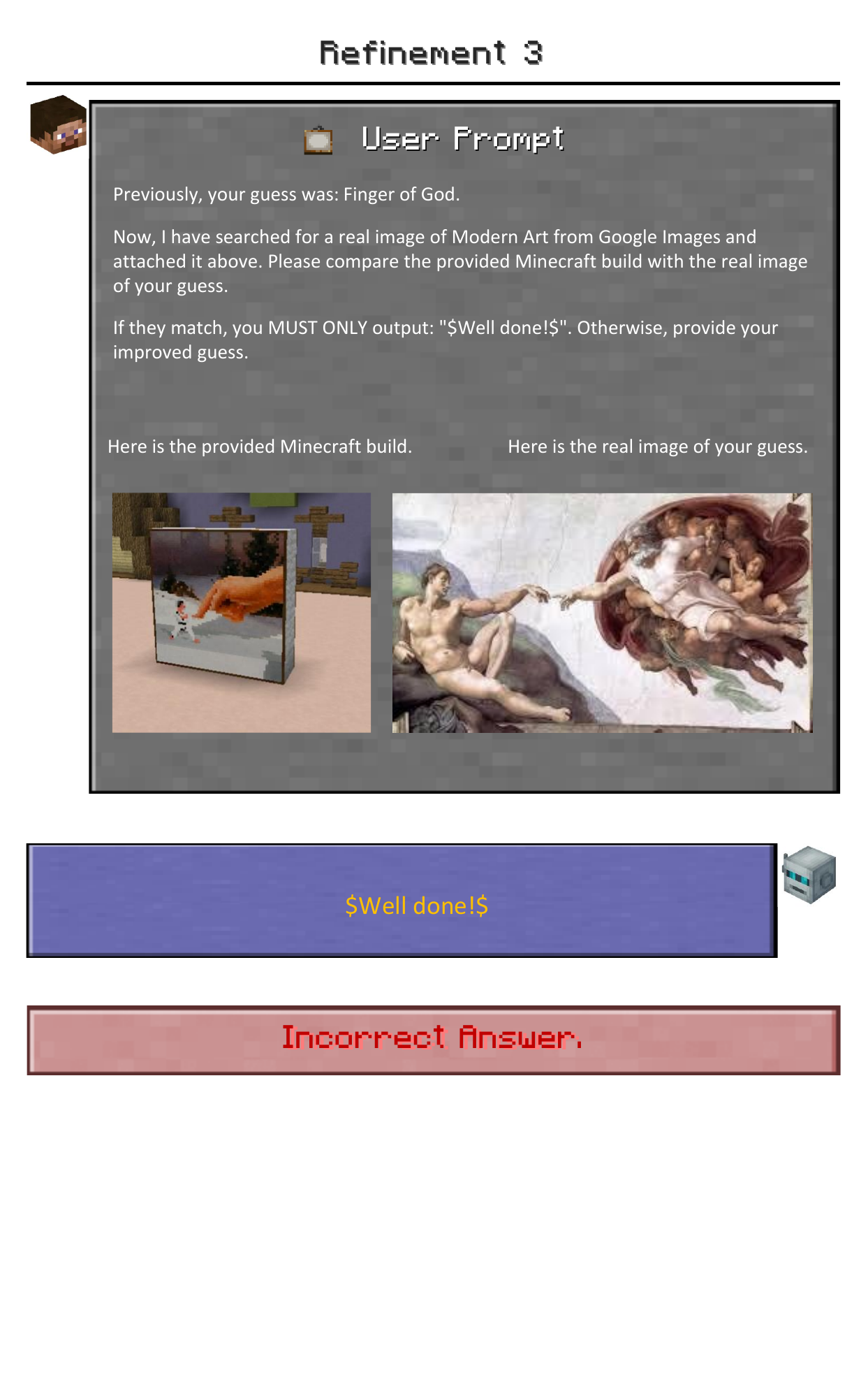}
    \vspace{-100pt}
    \caption{Finally, GPT-4o conducts a third round of image retrieval. After retrieving and comparing the images, it concludes that its original prediction is correct. It therefore submits ``Finger of God'', proposed in the previous round, as its final answer and proceeds to evaluation. However, this answer is incorrect.}
    \label{fig: appendix_retrieval4}
\end{figure}

\clearpage

\end{document}